\documentclass[10pt,twocolumn,letterpaper]{article}

\usepackage{cvpr}              % To produce the CAMERA-READY version
\usepackage[accsupp]{axessibility} % Improves PDF readability for those with disabilities.
\usepackage[dvipsnames]{xcolor}
\usepackage{sidecap}
\usepackage{multirow}
\usepackage{indentfirst}

\newcommand{\para}[1]{\vspace{.05in}\noindent\textbf{#1}}
\newcommand{\argmax}{\mathop{\mathrm{argmax}}}
\definecolor{cvprblue}{rgb}{0.21,0.49,0.74}
\usepackage[pagebackref,breaklinks,colorlinks,citecolor=cvprblue]{hyperref}

\def\paperID{13772} % *** Enter the Paper ID here
\def\confName{CVPR}
\def\confYear{2024}

\title{PDF: A Probability-Driven Framework for Open World 3D Point Cloud Semantic Segmentation}

\author{Jinfeng Xu, Siyuan Yang, Xianzhi Li \thanks{Corresponding author}, Yuan Tang, Yixue Hao, Long Hu, Min Chen\\
Huazhong University of Science and Technology\\
{\tt\small \{jinfengx, reedyoung, xzli, yuan\_tang, yixuehao, hulong, minchen2012\}@hust.edu.cn}
}

\begin{document}
\maketitle
\setlength{\parindent}{0.422 cm}
\begin{abstract}
% Existing point cloud semantic segmentation networks are strictly limited to closed-set and static scenarios, which cannot identify unknown classes and update their knowledge.
Existing point cloud semantic segmentation networks cannot identify unknown classes and update their knowledge, due to a closed-set and static perspective of the real world, which would induce the intelligent agent to make bad decisions.
%, which cannot identify unknown classes and update their knowledge.
To address this problem, we propose a \textbf{Probability-Driven Framework (PDF)}\footnote{Code available at: \href{https://github.com/JinfengX/PointCloudPDF}{https://github.com/JinfengX/PointCloudPDF}.}
for open world semantic segmentation that includes (i) a lightweight U-decoder branch to identify unknown classes by estimating the uncertainties, (ii) a flexible pseudo-labeling scheme to supply geometry features along with probability distribution features of unknown classes by generating pseudo labels, and (iii) an incremental knowledge distillation strategy to incorporate novel classes into the existing knowledge base gradually.
Our framework enables the model to behave like human beings, which could recognize unknown objects and incrementally learn them with the corresponding knowledge.
Experimental results on the S3DIS and ScanNetv2 datasets demonstrate that the proposed PDF outperforms other methods by a large margin in both important tasks of open world semantic segmentation.

\end{abstract}

\section{Introduction}
\label{sec:intro}

In recent years, deep learning for 3D point clouds has attracted increasing interest due to its great potential in various applications, such as virtual/augmented reality, robotics, autonomous driving, \etc.
Taking advantage of the emergence of high-quality datasets~\cite{Behley_2019_ICCV, Caesar_2020_CVPR, Armeni_2016_CVPR, Dai_2017_CVPR} and advances in point cloud networks~\cite{Qi_2017_CVPR, qi2017pointnet++, NEURIPS2022_d78ece66, Lai_2022_CVPR, wang2019dynamic}, the semantic segmentation task for point clouds has achieved promising performance in several important metrics.
Most existing methods work under the strong assumption that the world is \emph{closed-set} and \emph{static}, which supposes that all the object categories remain consistent in both the training and inference stages.
However, the assumption is not valid for many dynamic real-world scenarios in which \emph{unknown} object classes will be encountered outside the learning process.
% The intelligent agent could be confused under a closed-set design due to the incorrect classification of the unknown object classes.
Intelligent agents could make wrong decisions under a closed-set design due to incorrect recognition of unknown classes.
Moreover, the agent cannot update its knowledge base under a static perception of the world, while humans can continuously extend their learned knowledge without forgetting.
These problems limit the closed-set and static methods to particular scenarios.

\begin{figure}[t]
\centering
\includegraphics[width=\linewidth]{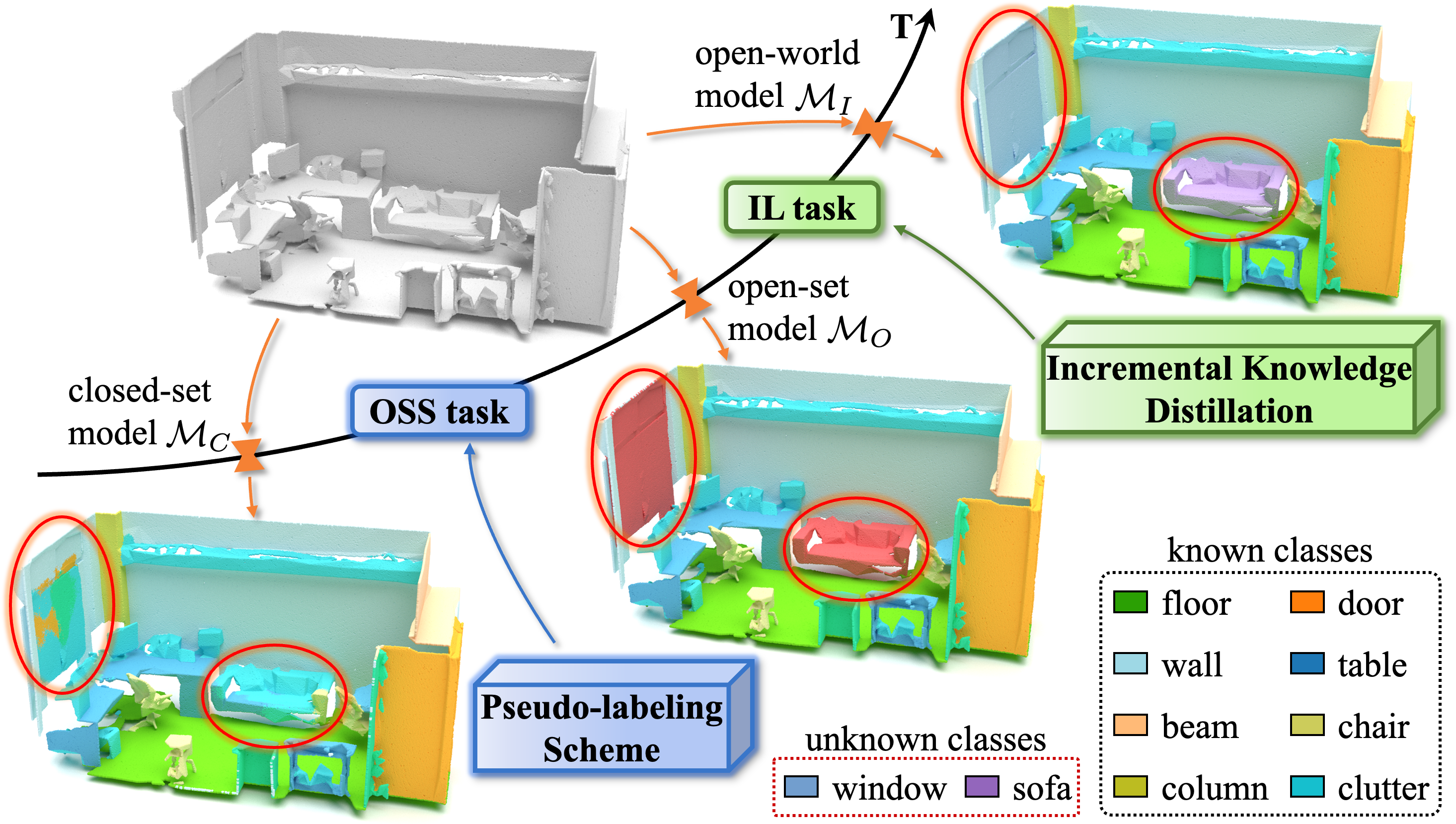}
\vspace{-1.2em}
\caption{
The closed-set model $\mathcal{M}_C$ continuously improves its open-world capabilities by successively finetuning to open-set model $\mathcal{M}_O$ and open-world $\mathcal{M}_I$ with the help of open-set semantic segmentation (OSS) task and incremental learning (IL) task, where the proposed pseudo-labeling scheme and incremental knowledge distillation strategy are employed, respectively.
}
\label{fig:teaser}
%\vspace{-0.8em}
\end{figure}

The open world semantic segmentation (OWSS) addresses the above issues by introducing two tasks: 1) open-set semantic segmentation (OSS) to recognize the known objects and identify unknown objects simultaneously; 2) incremental learning (IL) to update knowledge of the model without retraining from scratch when information about the identified unknown classes would be accessible.
Notably, the OSS focuses on identifying unknown classes not present during training, while OWSS addresses both unknown identification and continuous learning when novel classes with ground-truth labels are provided in IL task. Thus, OWSS can be considered as an extension of OSS.
% \red{Notably, the OSS focus on identifying unknown classes not present during training, while OWSS addresses both unknown identification and continuous learning when novel classes with ground-truth labels are provided in IL task. Thus, OWSS can be considered as an extension of OSS.}
\cref{fig:teaser} shows the OWSS pipeline for point clouds with an example, where unknown objects are marked with red circles.

Early works on the OSS task mainly focus on the 2D image domain.
Uncertainty estimation-based methods~\cite{hendrycks2017a, pmlr-v162-hendrycks22a, pmlr-v48-gal16} and generative network-based methods~\cite{10.1007/978-3-030-58452-8_9, Lis_2019_ICCV, Kong_2021_ICCV} are two main trends for the 2D OSS task.
% The former methods estimate the uncertainties of model outputs to identify the unknown objects.
% , which make a few modifications to the network.
% The latter methods locate the unknown objects by comparing the original inputs with the reconstructed inputs that are obtained from the conditional generative adversarial network.
% There are so few modifications to the network that uncertainty estimation-based methods rarely have a negative effect on the semantic segmentation performance.
% Thus, their techniques for OSS task can be extended to 3D OSS task with a tiny cost.
% However, uncertainty estimation-based methods usually have limited open-set capability in 3D scenes.
% The generative network-based methods are well-worked in 2D OSS, but struggle to reconstruct the dedicated geometry structures of the irregularly arranged point clouds.
% In addition, both methods cannot update their learned knowledge with novel classes. 
However, these methods suffer from performance loss when applied in 3D scenes.
% Although the OWSS problem has gained more attention with the goal of relaxing the closed-set and static assumption for realistic application, rarely studies have investigated in this problem in 3D scenes.
Although the OWSS problem has gained more attention for relaxing the assumption of realistic application, few studies have investigated it in 3D scenes.
Cen~\etal~\cite{10.1007/978-3-031-19839-7_19} first proposed a REAL framework to solve both OSS and IL tasks in a general architecture, making redundancy classifiers an important part of the framework.
% REAL introduced a calibration loss to uniformly force the probability distribution to the unknown classes.
% Meanwhile, REAL utilities the instance of known classes to synthesis the unknown objects by randomly resizing.
% REAL introduced a calibration loss and an unknown object synthesis scheme to make the features of unknown classes distinguishable from the perspective of probability and geometry.
% Very recently, 
Li~\etal~\cite{Li_2023_CVPR} proposed an Adversarial Prototype Framework (APF) to estimate the distribution of unknown classes with the help of learned prototypes.
% The APF achieved better performance on the 3D OSS task, while no further research has been conducted on the IL task.
Although REAL and APF have made satisfactory progress in the OSS task, they have drawbacks when applied to indoor 3D scenes, as discussed in~\cref{subsec:rw_owss}.

The present insufficient research situation motivates us to propose a novel OWSS framework, named the Probability-Driven Framework (PDF) for point clouds, which leverages the probability output (the last-layer output of the backbone) to solve both the OSS task and the IL task.
% As shown in~\cref{fig:teaser}, for the OSS task, we estimate the uncertainties of the unknown classes by designing a lightweight U-decoder, which is supervised with the pseudo labels that are generated by the pseudo-labeling scheme.
As shown in~\cref{fig:teaser}, for the OSS task, we identify the unknown objects by learning the geometry features and the implicit probability distribution features of the unknown classes with our proposed pseudo-labeling scheme.
% \textcolor{yellow}{The pseudo-labeling scheme utilizes the probability output to generate pseudo labels for supervising our designed lightweight U-decoder, which estimates uncertainties of unknown classes based on backbone features.}
% The pseudo labels locate the unknown objects in scenes to capture the geometry features along with implicit probability distribution features of unknown classes.
After that, the model incrementally expands its knowledge base through our proposed incremental knowledge distillation strategy for the IL task.
% After that, the \textcolor{green}{backbone(model)} incrementally expands its knowledge base through our proposed incremental knowledge distillation strategy for the IL task.
% , which transform the previous learned knowledge and the novel classes information in a distilled manner.
This process transforms previously acquired knowledge along with the information from novel classes in a distilled manner. 
% The effectiveness of above designs is verified in our experiments.
The quantitative and qualitative results show the superiority and effectiveness of our framework compared to the state of the arts.
% Notably, we explore the solution that leverages the probability output (the last layer's output) for both OSS and IL tasks.
% For OSS task, the pseudo-labeling scheme first locates the potential area of unknown classes by utilizing the probability distribution information, then builds an undirected graph based on the probability similarities to separate the unknown objects from context.
% with a heuristic unknown-aware (HUA) algorithm, then separate the objects from the area with a 3D graph boundary detection (GBD) algorithm.
% In this way, we capture the probability distribution features of the unknown classes as well as the geometry features of the unknown instance, which are beneficial to uncertainty estimation.
% The pseudo-labeling scheme 
% During the IL task, we obtain the pseudo probability labels through fusing the novel class labels and the probability output of the saved open-set model, which supplies the information about previously learned classes. 
% This would allow us to solve OWSS problem based on the probability output.% To valida te the effectiveness of our probability-based framework, we carried out experiments on two backbones~\cite{Zhao_2021_ICCV, Lai_2022_CVPR} across two widely used datasets~\cite{Armeni_2016_CVPR, Dai_2017_CVPR} to compared our method against other 
%  methods that are designed for OSS and IL tasks.
% The quantitative and qualitative results shows the superi-
% ority and effectiveness of our framework compared to state-of-
% the-art.
Overall, our contributions can be summarized as follows:
\begin{itemize}
    \item We propose a novel probability-driven framework (PDF) for open world semantic segmentation of point clouds.
    It requires less strict conditions in real-world applications.
    % \item We introduce a novel method for both OSS and IL task, based on a lightweight U-decoder and pseudo-labeling scheme and incremental knowledge distillation.
    % \item We introduce a novel pseudo-labeling scheme to capture the features for the unknown classes based on the probability output.
    % \item We propose a novel pseudo-labeling scheme which is designed for OSS task to capture features associated with unknown classes by leveraging probability outputs. 
    \item We propose a novel pseudo-labeling scheme designed for the OSS task to capture features of unknown classes by leveraging probability outputs.
    This approach enhances the model's ability to recognize unknown objects.
    % This approach enhances the representation of features pertaining to classes that may not be explicitly defined, thereby enriching the model's ability to recognize and classify a broader range of entities.
    % This approach enhances the feature of unknown objects.
    \item We propose a general incremental knowledge distillation strategy for the IL task to incorporate novel semantic classes into learned knowledge incrementally.
    % which estimate the uncertainties of unknown classes by learning the probability distribution and the geometric structure features of the unknown classes.
    % a lightweight U-decoder to address OSS task by estimating the uncertainties of unknown classes.
    % A novel pseudo-labeling scheme  for capturing the probability distribution and the geometric structure features of the unknown classes.
    % \item Our extensive experiments on two popular datasets demonstrate the effectiveness of the proposed PDF, which outperforms previous works significantly in both OSS task and IL task.
\end{itemize}

% To validate the effectiveness of our probability-based framework, we carried out experiments on two backbones~\cite{Zhao_2021_ICCV, Lai_2022_CVPR} across two widely used datasets~\cite{Armeni_2016_CVPR, Dai_2017_CVPR} to compare our method against other methods that are designed for OSS and IL tasks.
% The quantitative and qualitative results show the superiority and effectiveness of our framework compared to the state-of-the-art.

\section{Related Work}
\label{sec:rw}

\subsection{Closed-set 3D semantic segmentation}
% In recent years, deep learning for 3D semantic segmentation has attracted increasing interest from the research community.% for its powerful feature learning abilities.  
% The goal of closed-set semantic segmentation is to predict
% Numerous of approaches have been proposed for 3D semantic segmentation,
Recent 3D semantic segmentation methods can be broadly categorized into projection-based methods, voxel-based methods and point-based methods. 

Projection-based methods~\cite{Tatarchenko_2018_CVPR, Ando_2023_CVPR, xu2020squeezesegv3} use 2D CNNs to extract features from projected 3D scenes or shapes and then aggregate these features for label regression. 
% This pipline can reduce the complexity of 3D data. 
% However, these methods often suffer from incomplete or neglected 3D geometry and context information after projection.
% Naturally, incomplete 3D geometry and context information comes along with projection.
% To explicitly utilize geometry features, 
Voxel-based~\cite{Choy_2019_CVPR, Rethage_2018_ECCV, Dai_2018_CVPR, Meng_2019_ICCV, Riegler_2017_CVPR, Graham_2018_CVPR} methods employ 3D convolutions to predict semantic occupancy within a voxelized representation of the 3D scenes or shapes. %space on voxels, which encode a 3D scene or shape as a 3D volumetric grid.
Point-based methods~\cite{Hua_2018_CVPR, Wang_2018_CVPR, NEURIPS2018_f5f8590c, Wu_2019_CVPR, Zhao_2019_CVPR, NEURIPS2022_9318763d, Hu_2020_CVPR, Zhang_2020_CVPR, wang2019dynamic, Landrieu_2018_CVPR, Li_2019_ICCV, Lei_2019_CVPR} handle the segmentation task with efficient point cloud representation.
% Pioneering point-based methods, such as PointNet~\cite{Qi_2017_CVPR} and PointNet++~\cite{qi2017pointnet++}, employed Multi-Layer Perceptrons (MLPs) with symmetric functions to capture geometric features. 
Pioneering point-based methods~\cite{Qi_2017_CVPR,qi2017pointnet++} employed Multi-Layer Perceptrons (MLPs) with symmetric functions to capture geometric features. 
% Building upon their success, a series of point-based methods~\cite{Hua_2018_CVPR, Wang_2018_CVPR, NEURIPS2018_f5f8590c, Wu_2019_CVPR, Zhao_2019_CVPR, NEURIPS2022_9318763d, Hu_2020_CVPR, Zhang_2020_CVPR, wang2019dynamic, Landrieu_2018_CVPR, Li_2019_ICCV, Lei_2019_CVPR} are proposed for the end-to-end point cloud networks.
% for the direct performance of semantic segmentation on point cloud data.
% To further exploit local structures, some methods~\cite{wang2019dynamic, Landrieu_2018_CVPR, Li_2019_ICCV, Lei_2019_CVPR} established graph-based connectivity between points.
%2D computer vision and nature language processing, 
% With the popular of Transformers, which have demonstrated impressive deep learning capabilities, recent methods~\cite{NIPS2017_3f5ee243, Zhao_2021_ICCV, NEURIPS2022_d78ece66, Park_2022_CVPR, Lai_2022_CVPR, Lai_2023_CVPR} have used similar mechanisms to achieve improved segmentation accuracy.
With the popularity of Transformers~\cite{NIPS2017_3f5ee243}, recent methods~\cite{NEURIPS2022_d78ece66, Zhao_2021_ICCV, Park_2022_CVPR, Lai_2022_CVPR, Lai_2023_CVPR} have used similar mechanisms to improve segmentation accuracy.
%by using the similar mechanism, which can learn the context information with weight maps.

% Although the aforementioned methods have made remarkable progress in 3D semantic segmentation, they 
In particular, the aforementioned methods focused primarily on the closed-set and static settings, 
% where semantic labels remain consistent between training and inference scenarios.
However, the real world presents a dynamic environment in which novel objects of unknown classes may be encountered.
 % a variety of unseen objects with new class labels  can be encountered as the environment changes in the real world. 
Closed-set approaches have high confidence in recognizing known classes, but struggle to identify features of unknown classes.
Additionally, these approaches cannot update their models for novel classes.
Thus, these issues motivate us to design PDF to train an open-world model with the capability of identifying unknown objects and continuous learning.

\subsection{Open-set 2D semantic segmentation}
% The aim of 2D open-set semantic segmentation (OSS) is to identify both known and unknown classes within scenes.
% To adopt the closed-set methods to realistic application,
Early research of 2D open-set semantic segmentation (OSS) established robust baselines to estimate model uncertainties using maximum softmax probabilities (MSP)~\cite{hendrycks2017a} or maximum logit (MaxLogit)~\cite{pmlr-v162-hendrycks22a}.
MC-Dropout~\cite{pmlr-v48-gal16} and Ensembles~\cite{NIPS2017_9ef2ed4b} improved performance by approximating Bayesian inference, adopting a probabilistic perspective.
% Though these methods are reliable to detect incorrect predictions, they suffer from confusing samples.
% To enhance the distinction of misclassified data, 
Wang \etal~\cite{Wang_2021_ICCV} and Zhou \etal~\cite{Zhou_2021_CVPR} introduced redundancy classifiers to widen the gap between known and unknown classes.
Instead of quantifying uncertainties of unknown classes, the generative methods~\cite{10.1007/978-3-030-58452-8_9, Lis_2019_ICCV, Kong_2021_ICCV} locate unknown objects by comparing the original inputs against the reconstructed outputs of the generative models.
% The key idea behind these methods comes from the option that known classes could be recovered better than unknown classes.
% utilized generative models to  
% the original inputs and the reconstructed images,  known classes could be recovered better than unknown classes.
% Recently, utilizing generative models has became anthoer trends
% \textcolor{yellow}{However, generative methods suffer from inaccurate closed-set results.}
% It is worth noting that most above methods have not fully leveraged associations within classes.
% % the class features.
% To address this limitation, 
Hwang \etal~\cite{Hwang_2021_CVPR} proposed an exemplar-based approach for identifying novel classes by clustering.
Cen \etal~\cite{Cen_2021_ICCV} proposed to learn class prototypes using contrast clustering and then calculated the similarities of the features within the metric space.

% reference this in introduction section
Recent years have witnessed the rapid development of 2D OSS, and these methods have shown robustness, making them applicable to 3D OSS. 
% However, these methods could not handle the 3D OSS task well enough for the following potential reasons.
However, these methods could not handle the 3D OSS task well enough due to neglect of the special contextual information and an unbalanced semantic distribution in 3D scenes.
% For instance, with the support of adequate data, 2D backbones can capture more features of the known classes, which is beneficial for model inference. 
% 2D backbones can capture more features of the known classes with the support of adequate data, which is beneficial for model inference.
% inference in new scenes.
% Moreover, there are much more complex context information and unbalanced semantic distribution due to the additional depth dimension in 3D space.
% Moreover, 3D scenes introduce complex contextual information and an unbalanced semantic distribution.
% Unlike 2D images, where object boundaries can be clearly defined, it is challenging to separate objects from their context in a 3D space, thus limiting the application of the pseudo-labeling scheme~\cite{Gupta_2022_CVPR}.
% Moreover, 3D scenes introduce complex contextual information and an unbalanced semantic distribution.
% Unlike 2D images, where object boundaries can be clearly defined, it is challenging to separate objects from their context in a 3D space, thus limiting the application of the pseudo-labeling scheme~\cite{Gupta_2022_CVPR}.
% To address these issues, we propose a novel pseudo-labeling scheme for point clouds which locates the unknown objects by exploiting uncertainty estimation of the semantic output in a coarse to fine manner.

\subsection{Open world 3D semantic segmentation}
\label{subsec:rw_owss}
The open-world problem was first formulated in~\cite{Bendale_2015_CVPR}, which combines open-set recognition with incremental learning technology to extend the existing deep learning network to real-world settings.
As far as we know, rarely works have investigated the 3D open world semantic segmentation (OWSS). 
% Cen \etal~\cite{10.1007/978-3-031-19839-7_19} first proposed a novel framework called REAL to address the open-set semantic segmentation (OSS) task and the incremental learning (IL) task of OWSS for Lidar point clouds.
Cen \etal~\cite{10.1007/978-3-031-19839-7_19} first proposed a novel framework called REAL to address the OWSS task for Lidar point clouds.
REAL adds several redundancy classifiers (RCs) to the backbone to predict the probability of unknown classes.
The RCs are trained with a calibration loss to uniformly force the probability distribution to the unknown classes.
Meanwhile, REAL uses the instance of known classes to synthesize unknown objects by randomly resizing, which supplies geometric features for RCs.
% synthesizing unknown objects and calibrating the prediction distribution to handle open-set 3D semantic segmentation task.
% Then the model in the OSS task is saved for the IL task, where the original inputs are fed into the saved model for previously learned knowledge.
Then the model in the OSS task is saved to supply previously learned knowledge for the IL task.
% generating pseudo labels that are combined with labels of new introduced classes.
Very recently, Li \etal~\cite{Li_2023_CVPR} proposed an adversarial prototype framework (APF) that designs a feature adversarial module and a prototypical constraint module to aggregate the features of known and unknown classes.
% Very recently, Li \etal~\cite{Li_2023_CVPR} proposed an adversarial prototype framework (APF) that \textcolor{green}{designed(designs)} a feature adversarial module and a prototypical constraint module to aggregate the features of known and unknown classes.
The APF achieved better performance on the 3D OSS task, while no further research has been conducted on the IL task.
% REAL introduced a calibration loss to uniformly force the probability distribution to the unknown classes.
% Meanwhile, REAL utilities the instance of known classes to synthesis the unknown objects by randomly resizing.
% It is worth noting that both of the open-set methods tried to rearrange the distribution of features space so that enlarged the discrimination of the known and unknown classes.
% However, the geometry features and context information with the scenes of the unseen classes were ignored in both methods.

% Nevertheless, both REAL and APF got a performance loss when 
Both REAL and APF try to implicitly rearrange the distribution of features to enlarge the gap between the known classes and the unknown classes. 
However, the geometry structures of the unknown objects, as well as the associated probability distribution of the unknown classes, are ignored.
% Moreover, the indoor 3D scenes differ from the outdoor scenes in instance densities, number of background points and inter-class association, which makes the unknown object synthesis method of REAL brings less geometry information in indoor scenarios. 
% To address the 3D OWSS problem, we propose a novel pseudo-labeling scheme for OSS task to harvest both the explicit geometry features and implicit probability distribution features of the unknown classes
% % , which reveals the features of unknown objects. 
% In addition, we proposed an incremental learning strategy, called incremental knowledge distillation, to update the knowledge base along with novel classes by leveraging probability output.

\section{Open World Semantic Segmentation}
\label{sec:definition}

In this section, we formalize the definition and give the working pipeline of open world semantic segmentation (OWSS) in 3D point clouds.
At any time $t$, we assume that the set of known object classes $\mathcal{K}^t = \left\{ 1, 2, \cdots, C \right\} \subset \mathbb{N}^+$ is labeled in the training datasets.
In addition, there is a set of unknown classes $\mathcal{U}^t = \left \{ C+1, \cdots \right \}$ that may be encountered in the inference stage.
% Here, the point cloud $\mathbf{P}_i = \left\{ \mathbf{p}_1, \cdots, \mathbf{p}_N \right\}$ is composed of $N$ points $\mathbf{p}_i = \left( x_i, y_i, z_i, f_{i1}, \cdots, f_{ic} \right)$ and the label $\mathcal{Y}_i = \left\{ y_1, \cdots, \y_N \right\}$ 
% , where $\left( x_i, y_i, z_i \right)$ is the point coordinate and $\left(f_{i1}, \cdots, f_{ic} \right)$ is $c$ channels point features (e.g. point colors). 
% Let $\mathcal{D}^t = \left\{ \mathcal{P}^t, \mathcal{Y}^t \right\}$ be a dataset containing $M$ point clouds $\mathcal{P}^t = \left\{ \mathbf{P}_1, \cdots, \mathbf{P}_M,  \right\}$ with corresponding labels $\mathcal{Y}^t = \left\{ \mathbf{Y}_1, \cdots, \mathbf{Y}_M \right\}$.
% Each point clouds $\mathbf{P}_i = \left\{ \mathbf{p}_1, \cdots, \mathbf{P}_N \right\}$ are composed of $N$ points $\mathbf{p}_i = \left( x_i, y_i, z_i, f_{i1}, \cdots, f_{ic} \right)$, where $\left( x_i, y_i, z_i \right)$ is the point coordinate and $\left(f_{i1}, \cdots, f_{ic} \right)$ is $c$ channels point features (e.g. point colors). 
% For every point $\mathbf{p}_i$ in point clouds $\mathbf{P}_i$, 
% Each $\mathbf{Y}_i = \left\{ \mathbf{y}_1, \cdots, \mathbf{y}_N \right\} \in \mathcal{K}^t$ labels the point cloud $\mathbf{P}_i = \left\{ \mathbf{p}_1, \cdots, \mathbf{P}_N \right\}$
Each sample in the dataset is paired with a point cloud $\mathcal{P}_i = \left\{ \mathbf{p}_1, \cdots, \mathbf{p}_N \right\}$ and its associated label $\mathbf{Y}_i = \left\{ y_1, \cdots, y_N \right\}$, where $y_i \in \mathcal{K}^t$ is class label for point $\mathbf{p}_i$ and $N$ is the number of points.
Here, each point $\mathbf{p}_i$ is composed of a point coordinate $\left( x_i, y_i, z_i \right)$ and $c$ channels features $\left( f_{i1}, \cdots, f_{ic} \right)$.

As discussed in~\cref{sec:intro}, a closed-set model $\mathcal{M}_C$ trained with known classes $\mathcal{K}^t$ fails to recognize unknown object classes $\mathcal{U}^t$ under the OWSS condition.
However, the closed-set model $\mathcal{M}_C$ can be adapted to open-world applications by introducing the open-set semantic segmentation (OSS) task and incremental learning (IL) task.
During the OSS task, the model $\mathcal{M}_C$ will be finetuned to open-set model $\mathcal{M}_O$, which predicts the label of the points of the known classes $\mathcal{K}^t$ and identifies the points belonging to any of the unknown classes $\mathcal{U}^t$.
For the IL task, the points of identified unknown objects are annotated with novel class labels $\mathcal{K}_n = \left\{ C+1, \cdots, C+n \right\} \subset \mathcal{U}$, where $n$ is the number of novel classes.
% the model $\mathcal{M}_o$ is incrementally finetuned to the model $\mathcal{M}_i$ with novel class labels $\mathcal{K}_n$, which are forwarded from the identified unknown object points.
% Then model $\mathcal{M}_o$ 
% Then the point labels of unknown classes $\mathbf{U}^t \subset \mathcal{U}$ will be given for IL task.
% Note that, when 
% outputs falls into the known classes $\mathcal{K}^t$.
% While an open-world 3D semantic segmentation (OW3DSS) model $\mathcal{M}^t$ at time $t$ is trained to predict the point labels of the known classes $\mathcal{K}^t$ and further to identify points belonging to any of the unknown classes $\mathcal{U}^t$.
% However, model $\mathcal{M}_c$ can be finetuned to model $\mathcal{M}^t$ by introducing two OW3DSS task: open-set semantic segmentation (OSS) and incremental learning (IL).
Then the model $\mathcal{M}_O$ is incrementally finetuned to the model $\mathcal{M}_I$ with $\mathcal{K}_n$ so that the knowledge base is enlarged to $\mathcal{K}^{t+1} = \mathcal{K}^t + \mathcal{K}_n$.
Considering privacy protection and computation limitation in the open-world setting, the model $\mathcal{M}_I$ can only access $\mathcal{K}_n$ during its training process.
Note that the model $\mathcal{M}_I$ maintains the open-set semantic segmentation ability without forgetting all the previously learned knowledge.
This pipeline of OSS and IL task cycles in life-long time to continuously update the open-world semantic segmentation model.
% To further specify the above working pipeline of OWSS, a toy-example of model predictions is illustrated in the top row of~\cref{fig:architecture}.
% The closed-set model $\mathcal{M}_C$ assigns known class labels to the unseen objects (in red circles).
% In contrast, the model $\mathcal{M}_O$ and $\mathcal{M}_I$ can identify and segment the unseen objects respectively, which is more adaptability to real dynamic world.

% \section{Probability-based Pseudo-labeling Framework}
\section{Probability-Driven Framework}
\label{sec:method}

% \begin{figure*}[t]
% 	\centering
% 	\includegraphics[width=1\textwidth]{figures/Architecture.png}
%     \caption{Illustrating our proposed probability-driven framework (PDF).
    
%     }
% 	\label{fig:architecture}
% \end{figure*}

% \begin{figure*}[t]
% \centering
% \floatbox[{\capbeside\thisfloatsetup{capbesideposition={right,top}}}]{figure}[\FBwidth]
% {\caption{A test figure with its caption side by side}\label{fig:architecture}}
% {\includegraphics[width=0.75\textwidth]{figures/Architecture.png}}
% \end{figure*}

% \begin{figure*}[t]
%   \floatbox[{\capbeside\captionsetup[capbesidefigure]{labelsep=newline,justification=rightlast}\thisfloatsetup{capbesideposition={right,top}}}]{figure}[\FBwidth]
%   {\caption{aaaaaaaaaaaaaaaaaaaaaaaaaaaaaaaa}}
%   {
%     \includegraphics[width=0.8\textwidth]{figures/Architecture.png}
%   }
% \end{figure*}

\begin{SCfigure*}
  \centering
  \caption{
  (a) \textbf{Architecture of the probability-driven framework.}
  Given an input point cloud $\mathcal{P}_{in}$, the architecture produces semantic results $\mathcal{O}_S$ and uncertainty results $\mathcal{O}_U$, which are respectively supervised by closed-set GT and pseudo GT in the open-set semantic segmentation task (marked in blue).
  During incremental learning task (marked in green), $\mathcal{O}_S$ is further supervised by distilled GT.
  (b)~Pipeline of incremental knowledge distillation.
  (c) Pipeline of pseudo-labeling scheme, which consumes $\mathcal{O}_S$ and outputs pseudo mask for unknown classes.
  % The $\mathcal{O}_S$ is supervised by closed-set GT in the open-set semantic segmentation task (marked in blue) and supervised by distilled GT in the incremental learning task (marked in green).
  % The $\mathcal{O}_U$ is concatenated with $\mathcal{O}_S$ and jointly supervised by pseudo GT, which combines the closed-set GT and pseudo mask.
  % $\mathcal{O}_S$ is used to generate pseudo labels by pseudo-labeling scheme.
  % In the incremental learning task (highlighted in green), $\mathcal{O}_S$ is .
  %\vspace{-0.8em}
  }
  \includegraphics[width=0.72\textwidth]{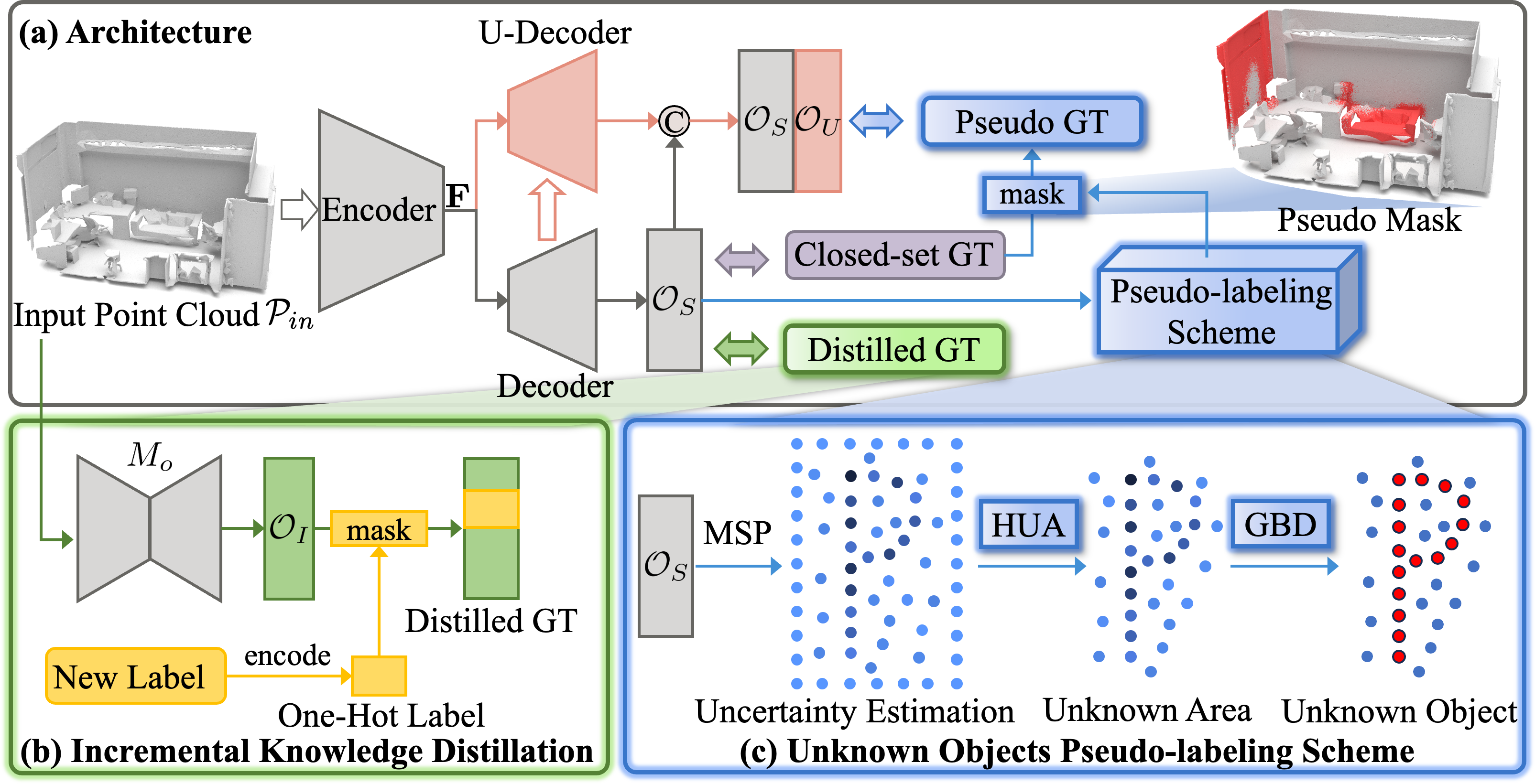}
  \label{fig:architecture}
%\vspace{-0.8em}
\end{SCfigure*}

% \begin{figure*}[t]
%   \begin{minipage}[c]{0.6\textwidth}
%     \includegraphics[width=\textwidth]{figures/Architecture.png}
%   \end{minipage}\hfill
%   \begin{minipage}[c]{0.3\textwidth}
%     \caption{
%        test caption.
%     } \label{fig:architecture}
%   \end{minipage}
% \end{figure*}

% \cref{fig:architecture} (a) shows the high-level overview of our proposed probability-driven framework (PDF) for open world 3D semantic segmentation (OW3DSS).
\Cref{fig:architecture} shows the high-level overview of our proposed probability-driven framework (PDF) for open world semantic segmentation (OWSS) of point clouds.
% The pipeline of our framework is composed of two stages, which addresses the open set semantic segmentation (OSS) task in the first stage (highlighted in blue color) and the incremental learning (IL) task in the second stage (highlighted in green color).
Our framework, which is composed of two stages, addresses the open set semantic segmentation (OSS) task in the first stage (highlighted in blue color) and the incremental learning (IL) task in the second stage (highlighted in green color).
% In the first stage, 
% We adopt the encoder-decoder structure as our backbone in both the open set semantic segmentation (OSS) task and the incremental learning (IL) task of OWSS.
The encoder-decoder structure is adopted as our backbone in both the OSS and IL tasks.
% For the OSS task, we propose (i) a lightweight U-Decoder branch to identify points of unknown classes by estimating the uncertainties of model outputs for each point; (ii) a flexible pseudo-labeling scheme which contains a heuristic unknown-aware (HUA) algorithm for searching potential areas of unknown objects in scenes and a 3D graph boundary detection (GBD) algorithm to further separate unknown objects from context (surrounding ground-truth known and background). 

For the OSS task, we first feed the raw point cloud $\mathcal{P}_{in} \in \mathbb{R}^{N \times (3+c)}$ with $N$ points and $c$ channel features into the backbone for the semantic segmentation output $\mathcal{O}_S \in \mathbb{R}^{N \times C}$, where $C \in \mathcal{K}^t$ is the number of known semantic classes.
Note that we treat $\mathcal{O}_S$ as the \textbf{ probability output } whose vectors could represent the semantic probabilities of the corresponding classes.
% element in the vectors represented the probability of  semantic class
Meanwhile, we send the backbone features to our designed lightweight U-decoder for the estimated uncertainties $\mathcal{O}_U \in \mathbb{R}^{N \times 1}$ of $\mathcal{O}_S$.
Then, we obtain pseudo masks through a pseudo-labeling scheme which will be detailed in~\cref{subsec:method_pseudo}
% which captures the features of unknown classes by exploiting the probability output $\mathcal{O}_S$.
% The semantic output $\mathcal{O}_S$ and is supervised by closed-set ground truth, which only contains the labels of the known classes.
The pseudo masks are used to generate the pseudo ground truth by masking the corresponding labels of known classes with $\max{\mathcal{K}^t}+1$.
% by fused with the labels of known classes by
% In this way, the pseudo ground truth that includes both known classes and unknown classes is generated.
Finally, the semantic output $\mathcal{O}_S$ is supervised by closed-set ground truth that only includes known classes.
For better performance, the uncertainty output $\mathcal{O}_U$ is concatenated with $\mathcal{O}_S$, then jointly supervised by the pseudo ground truth.
% , and concatenate $\mathcal{O}_U$ with $\mathcal{O}_S$.
% For the IL task, we design an incremental knowledge distillation strategy to preserve and update the learned knowledge base.
% For the IL task, we designed an incremental knowledge distillation strategy for preserving the learned knowledge in OSS task and updating the knowledge base with novel class information.
% In both OSS and IL task, we generate pseudo-label by leveraging the outputs of the model,
The trained model $\mathcal{M}_O$ will be saved for the following IL task.
% The trained model \red{$\mathcal{M}_O$} will be saved for the following IL task.
% During the IL task, we finetune the open-set model to open-world model with only the labels of $n$ novel classes $\mathcal{K}_n$.

During the IL task, only the labels of $n$ novel classes $\mathcal{K}_n$ are accessible.
% we finetune the open-set model to open-world model with only the labels of $n$ novel classes $\mathcal{K}_n = \left\{ C+1, \cdots, C+n \right\}$ to avoid retrain from scratch.
The number of backbone's head is correspondingly extended by $n$ to recognize the newly introduced classes.
The input point clouds are then fed into the saved open-set model $\mathcal{M}_O$ and open-world model $\mathcal{M}_I$ for output $\mathcal{O}_I$ and $O_S^{t+1}$, respectively.
% The input point clouds are then fed into the saved open-set model \red{$\mathcal{M}_O$} and open-world model \red{$\mathcal{M}_I$} for output $\mathcal{O}_I$ and $O_S^{t+1}$, \red{respectively}.
The output of the open-world model $O_S^{t+1}$ is supervised by distilled ground truth, which is generated by the proposed incremental knowledge distillation strategy (detailed in~\cref{subsec:method_il}) from $\mathcal{O}_I$ and novel class labels.
% The output of the open-world model $O_S^{t+1}$ is supervised by distilled ground truth, which is generated by the proposed incremental knowledge distillation strategy (\textcolor{yellow}{detailed in}~\cref{subsec:method_il}) from $\mathcal{O}_I$ and novel class labels.
Note that the open-set task is still in an active state to maintain the open-set capability when training the open-world model.

At the inference stage, the model prediction $\hat{\mathbf{Y}}$ is given based on the semantic segmentation output $\mathcal{O}_S$ and the estimated uncertainties $\mathcal{O}_U$ :
\begin{equation}
    \hat{\mathbf{Y}} = 
    \left\{ 
        \begin{aligned}
        & \argmax \mathcal{O}_S^i, && \mathcal{O}_U^i < \lambda \\
        & \max \mathcal{K}^t +1, && \mathcal{O}_U^i > \lambda,
        \end{aligned}
    \right.
\end{equation}
where $\lambda$ is the threshold to determine whether the $i$-th point belongs to unknown classes.
% Next, we present each component of the \red{proposed} PDF in detail.
Next, we present each component of the proposed PDF in detail.
% in~\cref{subsec:method_oss} and~\cref{subsec:method_il} respectively.
% To identify the points of unknown class for OSS task, we additional add a lightweight U-Decoder which consumes the encoder outputs and decoder features.
% \cref{fig:architecture} shows the overall architecture of the 
% In this section, we firstly propose the probabilistic inference pseudo-labeling (PIP) framework for open-world 3D semantic segmentation, which is composed of 
% Then 
% In this section, we first propose the probability-based pseudo-labeling (PPL) framework for open-world 3D semantic segmentation, which is composed of open-set semantic segmentation (OSS) and incremental learning (IL) tasks.
% Then we will detail the strategies in the OSS and IL tasks in \cref{subsec:method_oss} and \cref{subsec:method_il} respectively.
% The pipeline of open-world semantic segmentation can be divided into two consecutive phases.
% The first phase performs the open-set semantic segmentation task, which aims to segment 3D scenes and detect unknown objects simultaneously.
% % Then, one or more of the unknown objects are labeled for incremental learning task.
% The second phase conducts incremental learning with newly introduced labels from the detected unknown objects, then the pipline returns to the first phase.
% Similar to the human learning process, the open-world semantic segmentation model grows its abilities by introducing 

\subsection{Open-set 3D semantic segmentation (OSS)}
\label{subsec:method_oss}
Given a point cloud $\mathcal{P}_{in}$, the OSS task aims to train an open-set model $\mathcal{M}_O$ to predict per-point semantic labels from the probability output $\mathcal{O}_S$, as well as detect unknown objects by estimating the uncertainties $\mathcal{O}_U$ of the segmentation results $\mathcal{O}_S$.
\Cref{fig:architecture}~(a) illustrates the pipeline of the OSS task, $\mathcal{P}_{in}$ is first fed into an encoder to extract sparse features $\mathbf{F}$ with downsampling and embedding layers.
Then the features $\mathbf{F}$ are upsampled with a decoder to regress the probability output $\mathcal{O}_S$.
Closed-set labels $\mathbf{Y}_C$ are used to calculate closed-set semantic segmentation loss $\mathcal{L}_C$:
% which supplies information about known classes $\mathcal{K}^t$ 
% We employ  semantic segmentation loss of the known classes is formulated as:
\begin{equation}
    \mathcal{L}_C = CE ( \mathcal{O}_S, \mathbf{Y}_C ),
\end{equation}
where $CE(\cdot)$ is the cross entropy loss function.
% $\oplus$
% , which are supervised by the labels $\mathcal{Y} = \left\{ y_1, \cdots, y_N \right\}$ of known object classes $\mathcal{K}^t = \left\{ 1, 2, \cdots, C \right\}$, i.e., closed-set ground truth.
% Note that the closed-set GT $\mathcal{Y}$ does not contain information on unknown classes, leading to misclassified results in the unknown objects due to lower scores on the unknown class labels on the output logits $\mathcal{O}_S$.

With $\mathcal{O}_S$ in hand, some methods~\cite{hendrycks2017a,pmlr-v162-hendrycks22a,Cen_2021_ICCV,Li_2023_CVPR} calculate uncertainty scores for unknown classes with hand-crafted discrimination functions.
In contrast, we estimate the uncertainties of $\mathcal{O}_S$ using a lightweight U-decoder, which has a structure similar to the backbone decoder.
The U-decoder consumes the encoder features $\mathbf{F}$ as input and fuses the hidden layer output of the backbone decoder layers by skip connections, which can fully leverage the extracted geometric and scene-wise features of the backbone.
% A simple multilayer perceptron is used to get the U-decoder output $\mathcal{O}_U$, 
The output of the U-decoder $\mathcal{O}_U$ describes how confident the model is with its segmentation results $\mathcal{O}_S$.
Please refer to our supplementary material for detailed network architecture.
% Based on $\mathcal{O}_S$ and $\mathcal{O}_U$, the predicted open-set semantic segmentation result $\hat{\mathcal{Y}_i}$ of point $\mathbf{p}_i$ is given as:
% \begin{equation}
% \hat{\mathcal{Y}_i} = 
% \left\{ 
% \begin{aligned}
% & \argmax \mathcal{O}_S^i & & \mathcal{O}_U^i < \lambda \\
% & C+1 & & \mathcal{O}_U^i > \lambda
% \end{aligned}
% \right.
% \end{equation}
% where $\lambda$ is the threshold to determine the points of novel class.
% In this way, we decompose the OSS task into two parallel task: traditional semantic segmentation task and 
% To supervise the output of the U-decoder $\mathcal{O}_U$, we generate pseudo-ground truth by combining closed-set GT and our proposed pseudo-labeling scheme.
% To ensure a promise performance of the unknown classes identifying, we help U-decoder branch to learn the features of unknown classes.
% exploiting the probability output $\mathcal{O}_S$
To ensure the promising performance of the U-decoder branch, we generate pseudo labels with our proposed pseudo-labeling scheme (\cref{subsec:method_pseudo}) to indicate the features of unknown classes.
However, these pseudo labels are not used directly for supervision due to the potential risk that leads to an unbalanced distribution on $\mathcal{O}_U$.
We thus combined the pseudo labels with closed-set labels by mask operation to obtain the pseudo ground truth $\mathbf{Y}_P$.
% We convert the pseudo labels to masks for masking the corresponding closed-set labels with the new label $\max \mathcal{K}^t+1$.
% In this way, we obtain the pseudo ground truth $\mathbf{Y}_P$ that fuses the closed-set ground truth and the pseudo labels.
Then we employ the cross entropy loss on the concatenated U-decoder branch output $\mathcal{O}_U$ and the semantic output $\mathcal{O}_S$:
\begin{equation}
    \mathcal{L}_P = CE ( \mathcal{O}_S \oplus \mathcal{O}_U , \mathbf{Y}_P ),
    \label{eq:loss_pseudo}
\end{equation}
where $\oplus$ is the concatenation operation.
The overall open-set semantic segmentation loss $\mathcal{L}_O$ consists of both $\mathcal{L}_C$ and $\mathcal{L}_P$ , which are balanced with parameter $\alpha$:
\begin{equation}
    \mathcal{L}_O = \mathcal{L}_C + \alpha \mathcal{L}_P .
    \label{eq:loss_oss}
\end{equation}

\subsection{Pseudo-labeling scheme} 
\label{subsec:method_pseudo}
% As we introduced in~\cref{subsec:method_oss}, the U-decoder takes the backbone features as input and gives the estimated uncertainties $\mathcal{O}_U$ of the backbone outputs $\mathcal{O}_S$.
% However, the closed-set ground truth $\mathcal{Y}$ only gives the point labels of the known classes $\mathcal{K}^t$, which cannot be used to train the U-decoder branch.
% However, closed-set ground truth $\mathcal{Y}$ cannot be used directly to train the U-decoder.
The goal of the pseudo-labeling scheme is to generate pseudo labels of unknown classes by leveraging probability output $\mathcal{O}_S$ to supervise the uncertainty output $\mathcal{O}_U$ of U-decoder.
As shown in~\cref{fig:architecture}~(c), we first obtain the uncertainties of points to measure the uncertainties of objects.
% becase accurate pseudo labels relies on the uncertainty estimation of objects.
% Hendrycks~\etal~\cite{hendrycks2017a} found that a trained model tends to have a higher maximum softmax probability (MSP) in known examples than in unknown examples.
Hendrycks~\etal~\cite{hendrycks2017a} found that known examples tend to have a higher maximum softmax probability (MSP) than unknown examples.
% Inspired by MSP, 
Thus, we use the MSP to calculate the uncertainty score $\mathcal{S}$ of the $i$-th point $\mathbf{p}_i$:
% Thus, MSP can be taken as a reference for uncertainty scores.
% Based on the MSP, the uncertainty score $\mathcal{S}$ of the $i$-th point $\mathbf{p}_i$ is formulated as:
\begin{equation}
    \mathcal{S} \! \left(  \mathbf{p}_i \right) = \max\nolimits_k ( \exp{\mathcal{O}_S^{ik}} / \sum\nolimits_j \exp{\mathcal{O}_S^{ij}} ),
    \label{eq:uncertainty_score}
\end{equation}
where $\mathcal{O}_S^{ij}$ is the probability of $\mathbf{p}_i$ belonging to the $j$-th semantic class and $\max_k (\cdot) $ calculates the max value of the dimension where the index $k$ is located.
In particular, maximum logit (MaxLogit)~\cite{pmlr-v162-hendrycks22a} can also be used to calculate uncertainty scores.
% Naturally, a distinguishing distribution of the uncertainty scores between the known classes and the unknown ones would bring better performance in the task of identifying unknown objects. 
% Inspired by MSP, we design the pseudo-labeling scheme to generate reliable pseudo-labels for the U-decoder by leveraging the difference in distribution of uncertainty scores between the known classes and the unknown classes.
% The pseudo-labeling scheme is composed of a heuristic unknown-aware (HUA) algorithm and a 3D graph boundary detection (GBD) algorithm.
% As shown in~\cref{fig:architecture}~(c), the pseudo-labeling scheme first obtains the uncertainty estimation by calculating the MSP of $\mathcal{O}_S$.
Then, the HUA algorithm locates the unknown areas based on the per-point uncertainty scores.
The unknown areas usually cover both unknown objects and their surrounding points that belong to the background or other objects.
% The HUA algorithm iteratively searches the potential area of unknown classes with the help of segmentation outputs $\mathcal{O}_S$ and marks the points in these areas with pseudo labels.
% and marks these points with the index $\mathcal{I}_P$.
% The marked points usually cover both the unknown objects and their surrounding points that belong to background or other objects.
Thus, we further separate unknown objects from the other points by designing the GBD algorithm.
% , which is designed to harvest both the explicit geometry features and the implicit probability distribution features of the unknown classes based on the graph structure.
% , which connects the points with similarity relations.
% utilizes the minimum spanning tree (MST) to obtain the
% The detailed pipelines of the HUA algorithm and the GBD algorithm are provided in the following.

\para{Heuristic unknown-aware algorithm.} 
% The purpose of the heuristic unknown-aware (HUA) algorithm is to locate unknown areas containing objects of unknown classes by exploiting the information of probability output $\mathcal{O}_S$.
The HUA algorithm takes $\mathcal{O}_S$ as input and outputs pseudo labels that indicate the unknown area.
% selected points with index $\mathcal{I}_P$.
The key idea of HUA is that the uncertainty scores of known objects are probably higher than the uncertainty scores of unknown objects.
Therefore, unknown objects tend to be located in the area with lower uncertainty scores.
% Following this idea, we can cover the unknown objects by searching the unknown area according to the .
% It is worth noting that the 3D scenes vary a lot with diverse objects and geometry sizes.
% Directly crop the unknown areas with a fixed parameter for the uncertainty scores would bring extra bias. 
% To adapt the algorithm to more situations, we implement the UAH algorithm heuristically, which starts with a base condition and gradually forwards to the end conditions.
% \textcolor{red}{TODO:Whether to show the uncertainty score distribution of unknown points.}
In contrast to cropping the unknown areas with a fixed parameter, we start the unknown area search process with a set of seeds and then heuristically find the unknown area until the stop condition.

Specifically, assume that we already have the uncertainty scores $\mathcal{S}(\mathcal{O}_S) \in \mathbb{R}^{N \times 1}$ of all $N$ points from the probability output $\mathcal{O}_S$ by~\cref{eq:uncertainty_score}.
% , whose backbone is trained with the closed-set labels $\mathcal{Y}$.
We sort $\mathcal{S}(\mathcal{O}_S)$ in ascending order and select $m$ ($m \ll N$) seed points among the top $p$ percent of the sorted $\mathcal{S}(\mathcal{O}_S)$. 
% of all the $N$ points, where $n \ll N$
The selected seeds $\mathcal{P}^{0} = \left\{\mathbf{p}_1^0, \cdots, \mathbf{p}_m^0 \right\}$ got a promising low uncertainty score, which establishes the start conditions of the HUA algorithm.
Then, we iteratively merge more points into $\mathcal{P}^{0}$ to construct a point set with a low average uncertainty score, ensuring the incorporation of the majority of points associated with unknown classes into this set.
% until the majority of points belonging to unknown classes are included in this set.
% To achieve this goal, 
Let us start the iteration with $\mathcal{P}^{0}$.
We search for the $k$ nearest neighbors of $\mathcal{P}^{0}$ and measure the similarity between the seeds $\mathcal{P}^0$ and their neighbors $NN(\mathcal{P}^0) = \left\{nn(\mathbf{p}_{1}^0), nn(\mathbf{p}_{2}^0), \cdots, nn(\mathbf{p}_{m}^0) \right\}$ with the distance value and the uncertainty score.
Here, $nn(\mathbf{p}_{i}^0) = \left\{\mathbf{p}_{i1}^0, \mathbf{p}_{i2}^0, \cdots, \mathbf{p}_{ik}^0\right\} \in \mathbb{R}^{k \times 3}$ are the $k$ neighbor points of $\mathbf{p}_{i}^0$.
% and conditionally pick a part of them to . 
% to find points similar to the seeds.
% Specifically,
% $NN(\mathcal{P}^0) = \left\{\mathbf{p}_{11}^0, \mathbf{p}_{12}^0, \cdots, \mathbf{p}_{1k}^0, \cdots, \mathbf{p}_{n1}^0, \mathbf{p}_{n2}^0, \cdots, \mathbf{p}_{nk}^0 \right\}
% We use $L2$-norm to calculate the distance similarity scores $Sim_{dis}$ of $\mathcal{P}_s$ and  $\mathcal{P}_{nn}$:
The distance similarity matrix $Sim_{D}$ of $\mathcal{P}_s$ and  $NN(\mathcal{P}^0)$ are formulated as:
% \begin{small}
\begin{equation}
% Sim_{D} \! \left( \mathcal{P}^0, NN(\mathcal{P}^0) \right) \! = \! \left( \! \frac{ \Vert \mathbf{p}_i - nn(\mathbf{p}_i) \Vert^2 }{\max \Vert \mathbf{p}_i - nn(\mathbf{p}_i) \Vert^2} \! \right) \! \in \! \mathbb{F}^{ n \times k },
Sim_{D} = \left( \frac{ \Vert \mathbf{p}_i - nn(\mathbf{p}_i) \Vert^2 }{\max \Vert \mathbf{p}_i - nn(\mathbf{p}_i) \Vert^2} \! \right) \in \mathbb{R}^{ m \times k },
\label{eq:sim_dist}
\end{equation}
% \end{small}
where $\mathbf{p}_i \in \mathcal{P}^0$.
% To correctly reflect similarity of the uncertainty scores, we employ a negative exponent function to calculate the similarity matrix of the uncertainty scores $Sim_{U}$:
% The similarity matrix of the uncertainty scores $Sim_{U}$ is processed with a negative exponent function to correctly reflect the similarity of the uncertainty scores.
To accurately represent the similarity of uncertainty scores, we utilize a negative exponent function to calculate the uncertainty score similarity matrix $Sim_{U}$:
% \begin{small}
\begin{equation}
% Sim_{U} \! \left( \mathcal{P}^0, NN(\mathcal{P}^0) \right) \! = \! \exp{ \left( - \lvert S \! \left(  \mathbf{p}_i \right) \! - \! S \! \left( nn( \mathbf{p}_i ) \right) \rvert \right) } \! \in \! \mathbb{F}^{ n \times k }.
Sim_{U} = \exp{ \left( - \lvert \mathcal{S} \! \left(  \mathbf{p}_i \right) - \mathcal{S} \! \left( nn( \mathbf{p}_i ) \right) \rvert \right) } \in \mathbb{R}^{ m \times k }.
\label{eq:sim_uncertainty}
\end{equation}
% \end{small}
The overall similarity matrix is given as:
\begin{equation}
\label{eq:sim_total}
Sim \! \left( \mathcal{P}^0, NN(\mathcal{P}^0) \right) = Sim_{D} + Sim_{U}.
\end{equation}
% Similarity scores are used to extend the set of seed points $\mathcal{P}_s$.
The seed points $\mathcal{P}^0$ preferentially pick neighbors with high similarity from the matrix for updating.
% The closer the distance and probability value, the higher the similarity.
% Empirically, the top 50\% points $\mathcal{P}_{50}$ are selected with indexes $Sim \left( \mathcal{P}_s, \mathcal{P}_{nn} \right) \leq 50\% \times \max{Sim \left( \mathcal{P}_s, \mathcal{P}_{nn} \right)}$.
Empirically, we enlarge $\mathcal{P}^0$ to $\mathcal{P}^1 \in \mathbb{R}^{m^1 \times 3}$ ($m^1 > m$) with the points whose similarity values are in the top 50\% of $Sim \! \left( \mathcal{P}^0, NN(\mathcal{P}^0) \right)$.
% are selected for updating $\mathcal{P}^0$ to $\mathcal{P}^1 \in \mathbb{R}^{m^1 \times 3}$, where $m^1 > n$.
% Then, the algorithm updates $\mathcal{P}_s$ to $\mathcal{P}_s^1 = \mathcal{P}_s \cup \mathcal{P}_{50}$ and checks the stop condition, which determines whether to continue another iteration for searching similar neighbors of $\mathcal{P}_s^1$.
% Before starting another searching iteration, we check the stop condition, which controls the size of searched unknown areas.
% to determine whether to continue.
% The stop condition influences the number of final 
% points, which are selected as points of unknown objects.
Recall that the HUA algorithm starts from a small set of seed points $\mathcal{P}^0$ with almost the lowest uncertainty scores.
Thus, the average value of the uncertainty score of $\mathcal{P}^0$ will increase due to the heuristic search process.
% Thus, we set up the stop condtion to terminate the iteration by 
% Thus, we could leverage this information to set up the stop condition of the algorithm.
% use the average MSP value of the input points $\mathcal{P}_{in}$ as the algorithm stop conditions.
% Moreover, we add standard deviation of input points MSP to the stop condition with a hyperparameter $\lambda$, which can adjust the stop condition conveniently.
% We hope 
Generally, we expect the average uncertainty score of the $s$-th iteration output $\mathcal{P}^s$ to be below the average uncertainty score of the input points $\mathcal{P}_{in}$.
Based on the~\cref{eq:uncertainty_score}, the stop condition of the $s$-th iteration is given as:
% \begin{equation}
% \overline{Pb \left( \mathcal{P}_s^n \right)} < \overline{ Pb\left( \mathcal{P}_{in} \right)} - \lambda \cdot \sigma \left( Pb \left( \mathcal{P}_{in} \right) \right),
% \end{equation}
\begin{equation}
\overline{
\sum \mathcal{S} (\mathbf{p}_i)}
< 
\overline{ \sum \mathcal{S}(\mathbf{p}_{j})} 
- 
\lambda \cdot \sigma \! \left( \sum \mathcal{S} (\mathbf{p}_{j}) \right),
\end{equation}
where $\mathbf{p}_i \in \mathcal{P}^s$, $\mathbf{p}_j \in \mathcal{P}_{in}$, $\sigma (\cdot)$ is the standard deviation function and $\lambda$ is the hyperparameter to adjust the stop condition.
% The effect of $\lambda$ will be discussed in~\cref{subsec:ablation}.
Naturally, when $\lambda$ increases, the HUA has a more strict stop condition, \ie, the HUA output has lower uncertainty scores, which reduces the range of unknown areas.
% $\mathcal{P}^n$ is the updated seed points in the $n$-th iteration and $\mathcal{P}_{in}$ is the raw input point clouds.
% where $\mathcal{P}_s^n $ is the expanded seed points of $n$-th iteration and $\sigma$ is the standard deviation function.
% When $\lambda$ increases, the UAH algorithm has a more relaxed stop condition, so that the final results contain more points.
% In this way, the algorithm can adapt to scenes with multi unknown objects, which will be validated in~\cref{subsec:ablation}.

\para{3D graph boundary detection algorithm.}
% The UAH algorithm iteratively obtains a set of points with low MSP to generate pseudo labels. 
% The UAH algorithm iteratively obtains a set of points with low uncertainty to locate the unknown areas.
% The result inevitably includes unknown objects' surrounding points, which are incorrectly masked as points of unknown objects.
The unknown areas obtained by HUA include not only unknown objects but also other points of the background or known objects, which leads to pollution on the features of the unknown classes.
% The incorrectly masked points will pollute the probability distribution and geometry shape of pseudo labels. 
To address this issue, we design a 3D graph boundary detection (GBD) algorithm to separate the unknown objects from other surrounding points by detecting the boundaries of the unknown objects.
% Similarly to the 2D boundary detection algorithm, the GBD algorithm aims to separate real unknown objects from known objects or background points within the set of points $\mathcal{P}_s^n$ obtained by UAH.
% Unlike 2D images whose pixels are tightly and neatly arranged, point clouds are made up of a set of irregular points along with delicate geometry shapes and complex topology structures, making traditional boundary detection algorithm ineffectiveness.
Unlike 2D images, where pixels are tightly and neatly arranged, point clouds consist of irregularly positioned points with complex geometry structures, rendering traditional boundary detection algorithms ineffective.
% The object boundaries can be detected with just a special designed convolution kernel, which slides on the image to response the variation of color gradient.
% Unlike ordered 2D images, point clouds are made up of a set of irregular points, whose boundary information is hard to handle due to delicate geometry shapes and complex topology structures.
% Therefore, the traditional boundary detection algorithm is not applicable to point clouds.
% However, the idea of detecting gradient within adjacent pixels inspires us to develop the GBD algorithm, which builds a 3D ``image'' from the point cloud based on the graph structure. 
% However, based on the graph structure, we can build a 3D ``image'' by embedding the points with undirected edges into the graph.  
However, we could construct a 3D ``image" based on the graph structure by embedding points into an undirected graph and perform 3D boundary detection on this ``image'' by applying the idea of 2D boundary detection algorithm, which finds boundaries using the gradient produced from the difference between adjacent pixels. 
% , where the 3D boundary information is easier to handle.
% This approach allows us to detect the 3D object boundary with the idea of 2D boundary detection algorithm, which finds the boundary by leveraging the gradient produced from the difference between adjacent pixels.
% makes it easier to handle 3D boundary information.
% We find that the MSP of known objects is usually higher than 
% a and more concentrated distribution than the unknown objects.
% The difference of MSP distribution inspires us to design the GBD algorithm, which tries to divide the different MSP distribution by measuring the entropy of MSP distribution.

% Specifically, given the selected points $\mathcal{P}_s^n$ by UAH, we first calculate the similarity scores between each point of $\mathcal{P}_s^n$ and their neighbors $NN(\mathcal{P}_s^n)$ by~\cref{eq:sim_dist}, \cref{eq:sim_pb} and~\cref{eq:sim_total}.
Specifically, given the output of HUA algorithm $\mathcal{P}^s$, we first obtain the similarity value between $\mathcal{P}^s$ and their neighbors $NN(\mathcal{P}^s)$ by~\cref{eq:sim_dist}, \cref{eq:sim_uncertainty} and~\cref{eq:sim_total}.
Then we embed each point of $\mathcal{P}^s$ in an undirected graph $\mathcal{G}$ with the similarity matrix $Sim\left(\mathcal{P}^s, NN(\mathcal{P}^s)\right)$:
\begin{equation}
\label{eq:graph}
    \mathcal{G} = \left\{\mathcal{P}^s, Sim\left(\mathcal{P}^s, NN(\mathcal{P}^s) \right) \right\},
\end{equation}
where $\mathcal{P}^s$ and $Sim\left(\mathcal{P}^s, NN(\mathcal{P}^s)\right)$  are the nodes and the edge weights of $\mathcal{G}$ respectively. 
% When one node has a low similarity scores with adjacent nodes, the local area around this node has a   dispersed MSP distribution, i.e., this area has a high entropy.
% Thus, the entropy of the graph $\mathcal{G}$ can be measured by the sum of edge weights:
% \begin{equation}
%     \mathcal{E}(\mathcal{G}) = \sum Sim(\mathcal{P}_s^n)
% \end{equation}.
% In this way, we obtained an ordered 3D ``image'' $\mathcal{G}$, where the graph nodes are organized by graph edges.
% Note that, directly deal with $\mathcal{G}$ is inefficient, because each node is connected with dozens of neighbor nodes.
% in $\mathcal{G}$ so that 
% apply GBD algorithm is inefficient.
% To reduce the computational cost and algorithmic complexity, the graph $\mathcal{G}$ is further 
% croped with minimum spinning tree.
% To reduce computational cost and algorithmic complexity, we crop the redundancy edges by searching the minimum spanning tree (MST) $\mathcal{T}$ of $\mathcal{G}$.
Note that, directly dealing with $\mathcal{G}$ is inefficient, thus we crop the redundancy edges by searching the minimum spanning tree (MST) $\mathcal{T}$ of $\mathcal{G}$.
The MST connects all the nodes of $\mathcal{G}$ with minimum sum weights of existing edges, \ie, the nodes tend to connect with their least similar neighbors.
% Thus, the gradient variations are highlighted in the MST, which is beneficial for GBD algorithm.
% In this way, the 3D object boundary can be obtained with similar boundary detection algorithm of 2D image, which is aware of gradient variation in pixel features, such as color and brightness.
% As aforementioned in~\cref{subsec:method_pseudo}, the semantic output tends to have a higher MSP on known classes than unknown classes.
% Meanwhile, we found that the MSP distribution of known classes is more concentrated than the MSP distribution of unknown classes.

\begin{figure}[t]
\centering
\includegraphics[width=0.9\linewidth]{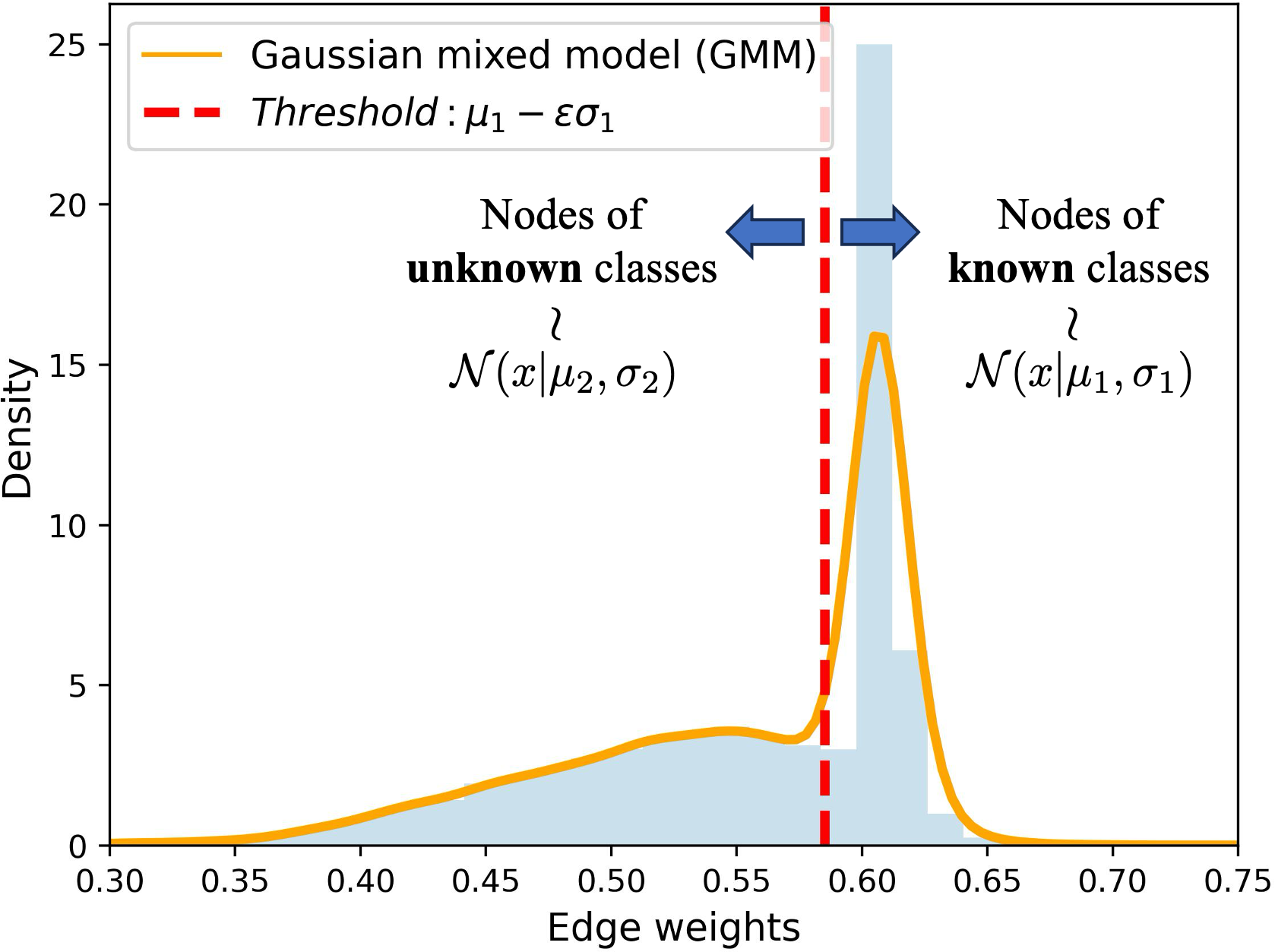}
\vspace{-0.5em}
\caption{
\textbf {Edge weights distribution.} The edges' weights between nodes of known classes are distinct from unknown classes. The distribution is approximately fitted with a Gaussian mixed model, and divided by the threshold $\mu_1 - \epsilon \sigma_1$ derived from the 3$\sigma$ criteria.
}
\label{fig:graph_edge}
\vspace{-0.8em}
\end{figure}

As aforementioned in~\cref{subsec:method_pseudo}, the unknown classes tend to have lower uncertainty scores than known classes.
Furthermore, we find that the uncertainty scores of the unknown classes are distributed in a wider region with lower values compared to the known classes.
% which is illustrated in~\cref{fig:graph_edge}.
% The difference in the distribution of the uncertainty scores is illustrated in~\cref{fig:graph_edge}, where the uncertainty scores of the unknown classes are distributed in a wider region with lower values compared to the known classes.
% If we use the uncertainty score as the node features, the gradient information of $\mathcal{T}$ is accessible.
% In addition, the difference of uncertainty scores between nodes will influence the graph edge weights according to~\cref{eq:sim_dist}, \cref{eq:sim_uncertainty} and \cref{eq:sim_total}, \ie, the connected nodes with higher similarity get a larger edge weights.
Thus, nodes of known classes probably get larger edge weights than nodes of unknown classes according to~\cref{eq:sim_dist}, \cref{eq:sim_uncertainty} and \cref{eq:sim_total}, as demonstrated in ~\cref{fig:graph_edge}.
% , \ie, the nodes of known classes tend to have higher edge weights.
% Thus, the gradient between nodes can be represented by the graph edge weights.
% Therefore, the nodes of known classes tend to have higher edge weights.
When the edges with high weights are cut off, most of the nodes belonging to known classes are isolated.
On the contrary, nodes belonging to unknown classes still preserve most of the edges, which are easy to distinguish from isolated nodes.
To achieve this goal, we approximately fit the overall distribution of the edge weights with a Gaussian Mixed Model (GMM):
% The overall distribution of the edge weights can be approximately fitted by 
\begin{equation}
    p(x) = \sum_{k=1}^{2} \pi_k \mathcal{N}(x | \mu_k, \sigma_k),
\end{equation}
where $\pi_k$, $\mu_k$ and $\sigma_k$ are the weight, mean and variance of the $k$-th Gaussian component respectively.
% We set $k=2$ to obtain the distribution of the unkown classes $\mathcal{N}(x | \mu_1, \sigma_1)$ and h
% \TODO simplify
Assuming that the distributions of the known classes and unknown classes are $\mathcal{N}(x | \mu_1, \sigma_1)$ and $\mathcal{N}(x | \mu_2, \sigma_2)$ respectively, we have $\mu_1 > \mu_2$ and $\sigma_1 < \sigma_2$.
Then, the edges whose weights are larger than $\mu_1 - \epsilon \sigma_1$ are cut off to split $\mathcal{T}$ into a group of subgraphs.
Here, $\mu_1 - \epsilon \sigma_1$ are the 3$\sigma$ outlier detection criteria.
% $\left\{ \mathbf{g}_1, \cdots, \mathbf{g}_m \right\}$.
% Due to the high edge weights among nodes of known classes, most edges of these nodes are cut off so that the subgraph of known classes contains much fewer points than the subgraph of unknown classes.
% Because of the high similarity between nodes belonging to known classes, most edges of these nodes are cut off so that 
% the subgraph of known classes contains much fewer points than unknown classes.
Finally, we obtain the unknown objects by merging the subgraphs whose number of nodes is not rejected by the outlier detection algorithm.

\begin{table*}[ht]
\captionsetup{skip=0.6em}
\centering
\caption{Open-set semantic segmentation results of 3D point clouds on S3DIS and ScanNetv2. We use the results in APF and mark these results with ``*''. The unavailable results are marked with ``-''. The best results are in bold in each metric.}
\label{tab:oss}\small
\resizebox{\textwidth}{!}{
    \begin{tabular}{l|ccc|ccc|ccc|ccc}
        \toprule
        \multirow{3}{*}{Methods} & \multicolumn{6}{c|}{S3DIS} & \multicolumn{6}{c}{ScanNetv2}\\
        \cmidrule(lr){2-7} \cmidrule(lr){8-13} & \multicolumn{3}{c|}{PointTransformer} & \multicolumn{3}{c|}{StratifiedTransformer} & \multicolumn{3}{c|}{PointTransformer} & \multicolumn{3}{c}{StratifiedTransformer} \\
         \cmidrule(lr){2-4} \cmidrule(lr){5-7} \cmidrule(lr){8-10} \cmidrule(lr){11-13} 
        & AUPR & AUROC & mIoU & AUPR & AUROC & mIoU & AUPR & AUROC & mIoU & AUPR & AUROC & mIoU \\
        \midrule
        MSP & 15.2* & 70.3* & \textbf{69.8}* & 16.7 & 72.4 & \textbf{70.5} & 35.6 & 75.0 & \textbf{64.5} & 36.8 & 77.8 & \textbf{64.9} \\
        MaxLogit & 17.5* & 74.3* & \textbf{69.8}* & 30.1 & 81.0 & \textbf{70.5} & 44.2 & 78.1 & \textbf{64.5} & 49.5 & 82.7 & \textbf{64.9} \\
        MC-Dropout & 18.2* & 75.9* & \textbf{69.8}* & 16.4 & 68.6 & 69.0 & 17.8 & 57.0 & 60.4 & 40.9&79.2&61.3 \\
        DMLNet & 20.5* & 80.7* & 67.2* & 17.4 & 66.3 & 70.4 & 36.6 & 76.8 & 63.5 & 30.3 & 76.1 & 61.6 \\
        REAL & 25.4* & 87.6* & 69.7* & 50.9& 87.1 & 70.4 & 42.4 & \textbf{87.3} & 63.9 & 58.1 & 90.3 & \textbf{64.9} \\
        APF & 31.6* & 90.0* & 69.3* & - & - & - & - & - & - & - & - & - \\
        \midrule
        \textbf{Ours} & \textbf{73.1} & \textbf{96.2} & 68.6 & \textbf{66.4} & \textbf{92.7} & 70.3 & \textbf{60.2} & 86.0 & 64.2 & \textbf{67.8} & \textbf{90.9} & 64.4 \\
        \bottomrule
    \end{tabular}
}
%\vspace{-0.8em}
\end{table*}

\subsection{Incremental learning (IL)}
\label{subsec:method_il}
IL task aims to finetune the open-set model $\mathcal{M}_O$ to the open-world model $\mathcal{M}_I$ by learning introduced novel class $\mathcal{K}_n$ without losing the open-set semantic segmentation ability.
Importantly, the training data only contains unknown classes at this stage to prevent retraining the model from scratch.
Finetuning the model with only the labels of $\mathcal{K}_n$ will lead to an erroneous prediction tendency for the novel classes, which is called catastrophic forgetting~\cite{french1999catastrophic}.
% We propose an incremental distillation loss, which integrate the labels of novel classes into the knowledge distillation process.
% Information that supplies the previous $C$ known classes can alleviate knowledge forgetting.
To alleviate knowledge forgetting, the open-set model $\mathcal{M}_O$ in open-set semantic segmentation (OSS) task is used as a teacher model to supply the knowledge of previously learned $C$ known classes for the IL task.
% The saved open-set model $\mathcal{M}_O$ contains the previous knowledge that is learned at open-set semantic segmentation (OSS) stage.
% Thus, the model $\mathcal{M}_o$ can play the role of teacher that helps student model $\mathcal{M}_i$ to maintain OSS performance in IL task.
Inspired by knowledge distillation~\cite{hinton2015distilling}, we transform knowledge from $\mathcal{M}_O$ to $\mathcal{M}_I$ by minimizing the distribution of probability output.

% The last layer output $\mathcal{O}_I \in \mathbb{R}^{N \times C}$ of the saved open-set model $\mathcal{M}_o$ is utilities to generate pseudo probability labels $\mathcal{Y}_P = \left\{ \mathbf{y}_{p}^1, \mathbf{y}_{p}^2, \cdots, \mathbf{y}_{p}^N \right\}$, where $N$ is the number of labels.
As shown in~\cref{fig:architecture} (b), the input point cloud $\mathcal{P}_{in}$ is sent into $\mathcal{M}_O$ and $\mathcal{M}_I$ for the probability output $\mathcal{O}_I$ and $\mathcal{O}_S^{t+1}$, respectively.
The output $\mathcal{O}_S^{t+1}$ is used to generate pseudo probability labels $\mathcal{Y}_I = \left\{ \mathbf{y}_1, \mathbf{y}_2, \cdots, \mathbf{y}_N \right\}$ for each input point, where $N$ is the number of points.
Each pseudo probability label $\mathbf{y}_i$ is obtained by distilling the corresponding soft target $ \mathbf{o}_I^i \in \mathcal{O}_I$:
\begin{equation}
\mathbf{y}_i = \mathcal{D} (\mathbf{o}_I^i,T) = \left[ \frac{\exp(\mathbf{o}_I^{ic}/T)}{ \sum_c \exp(\mathbf{o}_I^{ic}/T)} \right] \in \mathbb{R}^{1 \times C},
\end{equation}
% where $\mathbf{o}_I^{ic}$ is the semantic probability of $\mathbf{o}_I^i$ belonging to class $c$ and $T$ is the distillation temperature.
where $\mathbf{o}_I^{ic}$ is the probability of $\mathbf{o}_I^i$ belonging to the $c$-th semantic class and $T$ is the distillation temperature.
% Noteably, the $\mathcal{O}_I$ is the logistics output  pseudo probability labels further 
Then, we transform the labels of novel classes to one-hot format labels $\mathbf{E}_{I}$ for incorporating with $\mathcal{Y}_I$.
By masking $\mathcal{Y}_I$ with $\mathbf{E}_{I}$, we get the distilled ground truth $\mathcal{Y}_{D}$ :
\begin{equation}
\mathcal{Y}_{D} = 
\begin{cases}
\mathcal{Y}_I, & \text{if \ } \mathcal{Y}_{ \mathbf{p}_i } \not\in \mathcal{K}_n \\
\mathbf{E}_{I}, & \text{if \ } \mathcal{Y}_{ \mathbf{p}_i } \in \mathcal{K}_n,
\end{cases}
\end{equation}
where $\mathcal{Y}_{ \mathbf{p}_i }$ is the label of point $\mathbf{p}_i$.
% Similar to $\mathcal{Y}_P$, the semantic segmentation output $\mathcal{O}_S$ is distlled 
In this way, we obtain the previously learned knowledge and the new knowledge by combining the labels of novel classes with generated pseudo probability labels.
% The open-set model $\mathcal{M}_o$ 
% To alleviate knowledge forgetting, the open-set model $\mathcal{M}_o$ are used to 
% the last layer output $\mathcal{O}_I$ of open-set model $\mathcal{M}_o$ is used to generate pseudo probability labels.
% To save the previous learned knowledge, we generate pseudo probability labels by leveraging the saved open-set model $\mathcal{M}_o$
% The loss of the semantic segmentation branch output $\mathcal{O}_S$ of the incremental model $\mathcal{M}_i$ is computed by Kullback–Leibler divergence, which minimize the probability distribution of  are supervised by distilled ground truth $\mathcal{Y}_{D}$ with distillation loss $\mathcal{L}_distill ( \mathcal{O}_S, \mathcal{Y}_{D} )$:
The semantic segmentation loss of the IL task $\mathcal{L}_{I}$ is calculated by KL-divergence, which minimizes the probability distribution of $\mathcal{O}_S^{t+1}$ and $\mathcal{Y}_{D}$:
% The segmentaion output $\mathcal{O}_S$ of $\mathcal{M}_i$ 
% We employ KL-divergence to 
% The loss of semantic segmentation $\mathcal{L}_{distill}$ is obtained from the segmentaion output $\mathcal{O}_S$ and  by KL-divergence
\begin{equation}
    \mathcal{L}_{I} = \sum_{\mathbf{o}_i^{t+1}, \mathbf{y}_D^i} \mathcal{D}(\mathbf{o}_i^{t+1}) \log\left(\frac{ \mathcal{D} (\mathbf{o}_i^{t+1})}{\mathbf{y}_D^i}\right),
\end{equation}
where $\mathbf{o}_i^{t+1} \in \mathcal{O}_S^{t+1}$ and $\mathbf{y}_D^i \in \mathcal{Y}_{D}$.
% \begin{equation}
% \begin{align}
% L_{\text{distill}} &= \sum_x P(x) \log\left(\frac{P(x)}{Q(x)}\right) \\
% &= \sum_x P(x) \log P(x) - \sum_x P(x) \log Q(x) \\
% =&L_{\text{distill}} = -\sum_{i=1}^{n} (p_{\text{teacher}}^i \log(p_{\text{student}}^i)) \\
% =&L_{\text{distill}} = \frac{1}{2} \sum_{i=1}^{n} (y_{\text{teacher}}^i - y_{\text{student}}^i)^2 \\
% &\text{Softmax}(x)_i = \frac{e^{x_i}}{\sum_{j=1}^{n} e^{x_j}}
% \end{align}
% \end{equation}

\section{Experiments}
\label{sec:experiment}
\subsection{Datasets}
\label{subsec:experiment_dataset}

% \renewcommand{\para}[1]{\vspace{.05in}\noindent\textbf{#1}}
% We conducted experiments focusing on the Open World problem, which encompasses Open-set Semantic Segmentation (OSS) and Incremental Learning (IL) tasks. The experimental setups are detailed in Section \ref{subsec: set}, followed by OSS results in Section \ref{subsec: exp OSS}, and IL experiment discussions in Section \ref{subsec: exp IL}.
We conduct experiments on the S3DIS~\cite{Armeni_2016_CVPR} and ScanNetv2~\cite{Dai_2017_CVPR} datasets for both the open-set semantic segmentation (OSS) task and the incremental learning (IL) task of the open world semantic segmentation (OWSS) problem.
% The S3DIS dataset is a point cloud indoor scene dataset, which is captured from 271 rooms including 
S3DIS comprises 271 point cloud scenes, which are annotated with 13 semantic classes.
% ScanNetV2 is composed of 1,613 indoor scans, distributed as 1,201 for training, 312 for validation, and 100 for testing, each with point-wise semantic labels across 20 object categories.
ScanNetv2 includes 1613 indoor scans annotated with point-wise semantic labels in 20 categories.
For the OSS task, we follow the APF~\cite{Li_2023_CVPR} to select \{\emph{window}, \emph{sofa}\} as the unknown classes in S3DIS.
In addition, more classes including \{\emph{chair}, \emph{door}, \emph{refrigerator}, \emph{toilet}\} are labeled as unknown classes in ScanNetv2.
All unknown classes are annotated with the label ``-1".
For the IL task, the labels of the unknown classes are introduced with newly assigned labels to update the open-set model.

\subsection{Evaluation metrics}
\label{subsec:experiment_metrics}
The OSS is composed of closed-set segmentation task and unknown class identification task as discussed in~\cref{subsec:method_oss}.
We employ the mean class IoU (mIoU) for the former task.
The latter task adopts the area under the ROC curve (AUROC) and area under the precision-recall curve (AUPR) as metrics.
To evaluate the performance of IL, we employ mIoU for both previously learned known classes and newly introduced classes, which are denoted as mIoU\textsubscript{old} and mIoU\textsubscript{novel} respectively.

\subsection{Implementation details}
\label{subsec:experiment_implementation}
We apply our proposed probability-driven framework (PDF) on two backbones: PointTransformer~\cite{Zhao_2021_ICCV} and StratifiedTransformer~\cite{Lai_2022_CVPR}.
We implement the networks in PyTorch and trained on 4 Nvidia RTX 3090 GPUs for 3000 (on S3DIS) and 600 (on ScanNetv2) epochs with batch size of 16.
During IL task, we further finetune the network for about 200 epochs.
% \textcolor{yellow}{For PointTransformer, we use the SGD optimizer with initial learning rate, momentum, and weight decay set to 0.5, 0.9 and 0.0001, respectively.
% For StratifiedTransformer, we use the Adam optimizer, and the learning rate is initialized as 0.006.
% The learning rate is dropped by 10 times at 60\% and 80\% of the total steps.}
The SGD and Adam optimizer are used for PointTransformer and StratifiedTransformer, respectively.
The balance parameter $\alpha$ in~\cref{eq:loss_oss} is set to 0.001.
In S3DIS dataset, we respectively set $m$, $p$ and $\lambda$ as 20, 0.02 and 1.0 for the heuristic unknown-aware (HUA) algorithm in~\cref{subsec:method_pseudo}.
In the ScanNetv2 dataset, $m$, $p$ and $\lambda$ in HUA increase to 200, 0.15 and 2.0, respectively, due to a higher percentage of points belonging to unknown classes in scenes against S3DIS.

\subsection{Open-set semantic segmentation}
\label{subsec:experiment_oss}
We first compare our proposed PDF with all existing 3D OSS methods, including REAL~\cite{10.1007/978-3-031-19839-7_19} and APF~\cite{Li_2023_CVPR}.
In addition, we further adapt 2D OSS methods, such as MSP~\cite{hendrycks2017a}, MaxLogit~\cite{pmlr-v162-hendrycks22a}, MC-Dropout~\cite{pmlr-v48-gal16} and DMLNet~\cite{Cen_2021_ICCV}, to address 3D OSS tasks.
% \textcolor{yellow}{All methods adopt the same backbone for fair comparisons. }
For APF, the code is not released for now, we thus directly use the evaluation values published
in their original paper for comparison on S3DIS.
% In addition, the closed-set results of the backbones are reported as the upper bound for comparative closed-set evaluation.
% We first implement several state-of-the-art (SOTA) methods, encompassing a range of out-of-distribution (OOD) detection techniques including Maximum Softmax Probability (MSP), MaxLogit, and the uncertainty estimation method, Monte Carlo Dropout (MC-Dropout). % [MSP, MaxLogit, MC-Dropout]
% Further, we compared our approach with more advanced methods like Deep Metric Learning Network (DMLNet) and the recent REAL technique using redundancy classifiers.  % [Dmlnet, REAL]
% The corresponding results are reported in~\cref{tab:oss}. 
% Comprehensive results are presented in Table\ref{tab:OSS}.

\Cref{tab:oss} shows the comparison results.
MSP and MaxLogit achieve the best performance in the mIoU metric because there is no modification to the networks.
Our method obtained significantly better performance on the unknown class identification task across all experiments, with a slight sacrifice to the closed-set semantic segmentation results.
% Our method surpasses the other methods in identifying unknown classes in both datasets across two backbones, except the AUROC metric on ScanNetv2 with PointTransformer.
Note that we trained PointTransformer in the S3DIS dataset and got 69.2 on the mIoU metric, which is lower than the value supplied in APF.
DMLNet and MC-Dropout have slightly lower mIoU than other methods, which may be caused by the negative influence of modifications to the architecture and training process.
% The results presented above indicate that the proposed PDF is capable of achieving an improved balance between the closed-set segmentation task and the unknown class identification task performance.
The results presented above demonstrate that the proposed PDF is capable of enhancing the overall balance between closed-set segmentation and unknown class identification.

\begin{figure*}[t]
\centering
\includegraphics[width=\linewidth,scale=1.00]{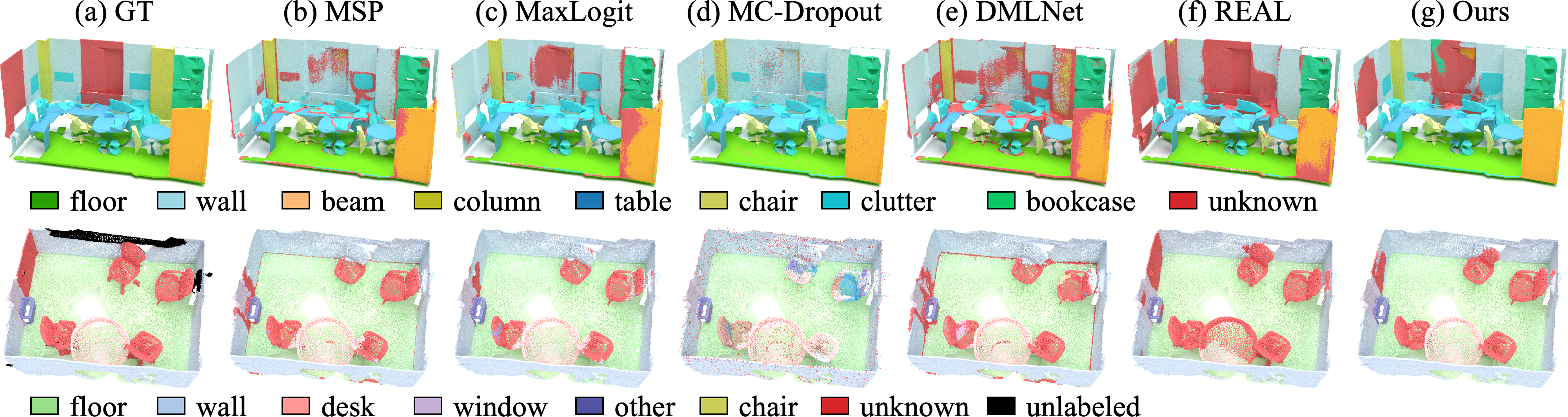}
\vspace{-1.0em}
\caption{Comparing the open-set semantic segmentation (OSS) results of our method (g) and other OSS methods (b-f).}
\label{fig:visual}
%\vspace{-0.8em}
\end{figure*}

\Cref{fig:visual} further shows the qualitative comparisons on two datasets, where the top scenes are from S3DIS
and the bottom scenes are from ScanNetv2. 
Clearly, the unknown objects identified by our method (g) are closest to the ground truths (a) against others (b-f).
More visual results are shown in our supplemental file.

\subsection{Incremental learning}
\label{subsec:experiment_il}
% In the Incremental Learning (IL) process, we introduced two tailored metrics: mIoU\textsubscript{old} and mIoU\textsubscript{novel}. These metrics are designed to specifically evaluate the model's performance on the originally trained $C$ classes (mIoU\textsubscript{old}) and the newly introduced $k$ classes (mIoU\textsubscript{novel}). This bifurcated approach allows for a nuanced assessment of the model's ability to retain knowledge of existing classes while effectively incorporating new ones, thereby providing a comprehensive evaluation of the IL procedure.

% Following the Incremental Learning (IL) method described in Sec.~\ref{sec:method_il}, we fine-tune the open-set model $\mathcal{M}_o$ to the incremental model $\mathcal{M}_i$. The model was pre-trained on datasets comprising $C$ classes and was then incrementally trained to recognize additional $n$ novel classes $\mathcal{K}_n$.

\begin{table}
\captionsetup{skip=0.5em}
\centering
\caption{Incremental learning results on S3DIS. There are 11 previously known semantic classes and 2 novel classes (window, sofa). The mIoU, mIoU\textsubscript{novel} and mIoU\textsubscript{old} are mean class IoU of all classes, previous known classes and novel classes, respectively.}
\label{tab:ILs3dis}\small
    \begin{tabular}{l|ccc}
        \toprule
        % \multicolumn{1}{c|}{S3DIS 11+2} & \multicolumn{2}{c|}{Validation set} \\
        % \cmidrule{1-5}
        \multicolumn{1}{c|}{Methods} & mIoU & mIoU\textsubscript{novel} & mIoU\textsubscript{old} \\
        \midrule
        Closed-set       & 58.6 & 0.0    &  69.2 \\
        Upper bound      & 70.1&72.0 &69.8 \\
        \midrule
        Finetune         & 0.5&3.0&0.0      \\
        Feature extraction & 57.3&9.5 &66.1 \\
        LwF              & 62.3&23.9 &69.2   \\
        REAL             & 68.9&61.3&70.3 \\
        \midrule
        \textbf{Ours}  & \textbf{69.4}&\textbf{64.3} &\textbf{70.3} \\
        \bottomrule
    \end{tabular}
%\vspace{-0.8em}
\end{table}

Since most OSS methods are not involved in the IL task, we thus carry out experiments on S3DIS with the methods designed for the IL task, \eg LwF~\cite{li2017learning} and Feature Extraction.
We further compare the proposed PDF with REAL, which is designed for OWSS, in terms of the IL task.
The result of directly finetuning the open set model $\mathcal{M}_O$ to the open-world model $\mathcal{M}_I$ with only novel classes is reported to illustrate catastrophic forgetting, as discussed in~\cref{subsec:method_il}.
The closed-set result is derived from the model trained with the known classes.
% In addition, the upper bound retrains the model with both the known and unknown classes, which is forward as oracle.
Furthermore, the upper bound retrains the model using both known and unknown classes, essentially treating it as an oracle during the forward process.
% \textcolor{yellow}{To mitigate network interference,} all approaches are implemented on the PointTransformer.
All approaches are implemented on the PointTransformer.

The results of the IL task are demonstrated in~\cref{tab:ILs3dis}.
Directly finetuning the open-set model using novel classes leads to serious bias, which incorrectly assigns the labels of the novel class to all points, \ie, causes catastrophic forgetting.
Our method achieves the best performance in both known classes and newly introduced classes compared to the others, showing the effectiveness of the incremental knowledge distillation strategy.
Compared with the upper bound, the knowledge of learned known classes is transferred to the open-world model without changing the networks, but experiences a performance degradation when it comes to learning new categories.
Further detailed results on the IL task are shown in our supplemental file. 

\subsection{Ablation study} 
To evaluate the individual contribution of each component in our framework, we perform ablation studies on the ScanNetv2 dataset, utilizing the StratifiedTransformer as the backbone.
We re-trained the network separately for following each experiment and reported the key results in~\cref{tab:ablation}.
Please refer to the supplementary material for more details.
% The purpose of these studies is to cut up the overall architecture by systematically removing or altering specific sub-modules and parameters, thereby isolating their individual impacts on performance and providing deeper insights into their functionalities and importance. 
% The quantified results of these studies are presented in Table \ref{tab:Ab}, offering a clear and factual basis for the comparative analysis that follows.
% \para{Unknown Prediction.}
% \para{Pseudo-label with real label.}
% \para{Hyperparameters }
% \para{Pseudo-label algo}
% \para{Soft \& hard target in IL}
\begin{itemize}
    \item Model A replaces the U-decoder with an MLP layer.
    \item Model B removes the semantic output $\mathcal{O}_S$ in~\cref{eq:loss_pseudo}.
    \item Model C removes the pseudo-labeling and loss $\mathcal{L}_P$.
    \item Model D removes the GBD algorithm in~\cref{subsec:method_pseudo}.
\end{itemize}
By comparing model C \& D \vs our full pipeline, we can see that both HUA and GDB in the pseudo-labeling scheme contribute to a better performance for open-set semantic segmentation.
In particular, the GDB algorithm can enhance both closed-set and open-set abilities, which can be seen by comparing D with ours.
By comparing model A with our full pipeline, we can see that the designed U-decoder can improve task performance.
By comparing models A \& B, we find that the open-set task has a serious setback when uncertainty output $\mathcal{O}_U$ is not supervised together with probability output $\mathcal{O}_S$, which may be caused by the imbalanced distribution in the supervisory signal.

\label{subsec:ablation}
\begin{table}
\captionsetup{skip=0.5em}
\centering
\caption{Comparing the contribution of major components
in our framework on ScanNetv2 using StratifiedTransformer.}
\label{tab:ablation}\small
    \begin{tabular}{l|ccc}
        \toprule
        \multicolumn{1}{c|}{Experiment} & AUPR & AUROC & mIoU \\
        \midrule
        A      & 67.6 & 90.6 &  64.1 \\
        B      & 31.0 & 76.0 & 64.2 \\
        C      & 20.0 & 62.0 & \textbf{64.9} \\
        D      & 64.3 & 85.8 &63.0 \\
        \midrule
        Ours & \textbf{67.8} & \textbf{90.9} & 64.4 \\
        \bottomrule
    \end{tabular}
%\vspace{-0.8em}
\end{table}

\section{Conclusion}
\label{sec:conclusion}

In this work, we present a novel Probability-Driven Framework (PDF) for open world semantic segmentation of 3D point clouds. Our framework addresses both open-set semantic segmentation and incremental learning tasks, by designing a lightweight U-decoder for estimating uncertainties of unknown classes, a pseudo-labeling scheme to generate ground truth for unknown classes, and an incremental knowledge distillation strategy for integrating novel classes into the existing knowledge base.
Both quantitative and qualitative results demonstrate that our framework significantly outperforms the state-of-the-art.
% \red{Nevertheless, our method's performance is constrained to capturing the geometry and probability distribution of objects in outdoor scenes where sparse-occupied and severely incomplete objects are frequently misrecognized with high confidence.}
Nevertheless, our method's performance is constrained to capturing the geometry and probability distribution of objects in outdoor scenes, where sparse-occupied and severely incomplete objects are frequently misclassified with high confidence.
In the future, we shall explore the possibility of utilizing more network features to improve task performance.
We hope that our work can attract more attention to this practically significant open problem.

\section*{Acknowledgments}
% \noindent This work is supported by the China National Natural Science Foundation No\text{.} 12345678, No\text{.} 12345678, No\text{.} 12345678.
This work is supported by the National Natural Science Foundation of China (NSFC) No.62202182, No.62176101, No.62276109, No.62322205.

{
    \small
    \bibliographystyle{ieeenat_fullname}
    \bibliography{main}

\begin{thebibliography}{50}
\providecommand{\natexlab}[1]{#1}
\providecommand{\url}[1]{\texttt{#1}}
\expandafter\ifx\csname urlstyle\endcsname\relax
  \providecommand{\doi}[1]{doi: #1}\else
  \providecommand{\doi}{doi: \begingroup \urlstyle{rm}\Url}\fi

\bibitem[Ando et~al.(2023)Ando, Gidaris, Bursuc, Puy, Boulch, and Marlet]{Ando_2023_CVPR}
Angelika Ando, Spyros Gidaris, Andrei Bursuc, Gilles Puy, Alexandre Boulch, and Renaud Marlet.
\newblock Rangevit: Towards vision transformers for 3d semantic segmentation in autonomous driving.
\newblock In \emph{CVPR}, pages 5240--5250, 2023.

\bibitem[Armeni et~al.(2016)Armeni, Sener, Zamir, Jiang, Brilakis, Fischer, and Savarese]{Armeni_2016_CVPR}
Iro Armeni, Ozan Sener, Amir~R. Zamir, Helen Jiang, Ioannis Brilakis, Martin Fischer, and Silvio Savarese.
\newblock 3d semantic parsing of large-scale indoor spaces.
\newblock In \emph{CVPR}, 2016.

\bibitem[Behley et~al.(2019)Behley, Garbade, Milioto, Quenzel, Behnke, Stachniss, and Gall]{Behley_2019_ICCV}
Jens Behley, Martin Garbade, Andres Milioto, Jan Quenzel, Sven Behnke, Cyrill Stachniss, and Jurgen Gall.
\newblock Semantickitti: A dataset for semantic scene understanding of lidar sequences.
\newblock In \emph{ICCV}, 2019.

\bibitem[Bendale and Boult(2015)]{Bendale_2015_CVPR}
Abhijit Bendale and Terrance Boult.
\newblock Towards open world recognition.
\newblock In \emph{CVPR}, 2015.

\bibitem[Caesar et~al.(2020)Caesar, Bankiti, Lang, Vora, Liong, Xu, Krishnan, Pan, Baldan, and Beijbom]{Caesar_2020_CVPR}
Holger Caesar, Varun Bankiti, Alex~H. Lang, Sourabh Vora, Venice~Erin Liong, Qiang Xu, Anush Krishnan, Yu Pan, Giancarlo Baldan, and Oscar Beijbom.
\newblock nuscenes: A multimodal dataset for autonomous driving.
\newblock In \emph{CVPR}, 2020.

\bibitem[Cen et~al.(2021)Cen, Yun, Cai, Wang, and Liu]{Cen_2021_ICCV}
Jun Cen, Peng Yun, Junhao Cai, Michael~Yu Wang, and Ming Liu.
\newblock Deep metric learning for open world semantic segmentation.
\newblock In \emph{ICCV}, pages 15333--15342, 2021.

\bibitem[Cen et~al.(2022)Cen, Yun, Zhang, Cai, Luan, Tang, Liu, and Yu~Wang]{10.1007/978-3-031-19839-7_19}
Jun Cen, Peng Yun, Shiwei Zhang, Junhao Cai, Di Luan, Mingqian Tang, Ming Liu, and Michael Yu~Wang.
\newblock Open-world semantic segmentation for lidar point clouds.
\newblock In \emph{European Conference on Computer Vision}, pages 318--334. Springer, 2022.

\bibitem[Choy et~al.(2019)Choy, Gwak, and Savarese]{Choy_2019_CVPR}
Christopher Choy, JunYoung Gwak, and Silvio Savarese.
\newblock 4d spatio-temporal convnets: Minkowski convolutional neural networks.
\newblock In \emph{CVPR}, 2019.

\bibitem[Dai et~al.(2017)Dai, Chang, Savva, Halber, Funkhouser, and Niessner]{Dai_2017_CVPR}
Angela Dai, Angel~X. Chang, Manolis Savva, Maciej Halber, Thomas Funkhouser, and Matthias Niessner.
\newblock Scannet: Richly-annotated 3d reconstructions of indoor scenes.
\newblock In \emph{CVPR}, 2017.

\bibitem[Dai et~al.(2018)Dai, Ritchie, Bokeloh, Reed, Sturm, and Nießner]{Dai_2018_CVPR}
Angela Dai, Daniel Ritchie, Martin Bokeloh, Scott Reed, Jürgen Sturm, and Matthias Nießner.
\newblock Scancomplete: Large-scale scene completion and semantic segmentation for 3d scans.
\newblock In \emph{CVPR}, 2018.

\bibitem[French(1999)]{french1999catastrophic}
Robert~M French.
\newblock Catastrophic forgetting in connectionist networks.
\newblock \emph{Trends in Cognitive Sciences}, 3\penalty0 (4):\penalty0 128--135, 1999.

\bibitem[Gal and Ghahramani(2016)]{pmlr-v48-gal16}
Yarin Gal and Zoubin Ghahramani.
\newblock Dropout as a bayesian approximation: Representing model uncertainty in deep learning.
\newblock In \emph{Proceedings of The 33rd International Conference on Machine Learning}, pages 1050--1059, New York, New York, USA, 2016. PMLR.

\bibitem[Graham et~al.(2018)Graham, Engelcke, and van~der Maaten]{Graham_2018_CVPR}
Benjamin Graham, Martin Engelcke, and Laurens van~der Maaten.
\newblock 3d semantic segmentation with submanifold sparse convolutional networks.
\newblock In \emph{CVPR}, 2018.

\bibitem[Hendrycks and Gimpel(2017)]{hendrycks2017a}
Dan Hendrycks and Kevin Gimpel.
\newblock A baseline for detecting misclassified and out-of-distribution examples in neural networks.
\newblock In \emph{ICLR}, 2017.

\bibitem[Hendrycks et~al.(2022)Hendrycks, Basart, Mazeika, Zou, Kwon, Mostajabi, Steinhardt, and Song]{pmlr-v162-hendrycks22a}
Dan Hendrycks, Steven Basart, Mantas Mazeika, Andy Zou, Joseph Kwon, Mohammadreza Mostajabi, Jacob Steinhardt, and Dawn Song.
\newblock Scaling out-of-distribution detection for real-world settings.
\newblock In \emph{Proceedings of the 39th International Conference on Machine Learning}, pages 8759--8773. PMLR, 2022.

\bibitem[Hinton et~al.(2015)Hinton, Vinyals, and Dean]{hinton2015distilling}
Geoffrey Hinton, Oriol Vinyals, and Jeff Dean.
\newblock Distilling the knowledge in a neural network.
\newblock \emph{arXiv preprint arXiv:1503.02531}, 2015.

\bibitem[Hu et~al.(2020)Hu, Yang, Xie, Rosa, Guo, Wang, Trigoni, and Markham]{Hu_2020_CVPR}
Qingyong Hu, Bo Yang, Linhai Xie, Stefano Rosa, Yulan Guo, Zhihua Wang, Niki Trigoni, and Andrew Markham.
\newblock Randla-net: Efficient semantic segmentation of large-scale point clouds.
\newblock In \emph{CVPR}, 2020.

\bibitem[Hua et~al.(2018)Hua, Tran, and Yeung]{Hua_2018_CVPR}
Binh-Son Hua, Minh-Khoi Tran, and Sai-Kit Yeung.
\newblock Pointwise convolutional neural networks.
\newblock In \emph{CVPR}, 2018.

\bibitem[Hwang et~al.(2021)Hwang, Oh, Lee, and Han]{Hwang_2021_CVPR}
Jaedong Hwang, Seoung~Wug Oh, Joon-Young Lee, and Bohyung Han.
\newblock Exemplar-based open-set panoptic segmentation network.
\newblock In \emph{CVPR}, pages 1175--1184, 2021.

\bibitem[Kong and Ramanan(2021)]{Kong_2021_ICCV}
Shu Kong and Deva Ramanan.
\newblock Opengan: Open-set recognition via open data generation.
\newblock In \emph{ICCV}, pages 813--822, 2021.

\bibitem[Lai et~al.(2022)Lai, Liu, Jiang, Wang, Zhao, Liu, Qi, and Jia]{Lai_2022_CVPR}
Xin Lai, Jianhui Liu, Li Jiang, Liwei Wang, Hengshuang Zhao, Shu Liu, Xiaojuan Qi, and Jiaya Jia.
\newblock Stratified transformer for 3d point cloud segmentation.
\newblock In \emph{CVPR}, pages 8500--8509, 2022.

\bibitem[Lai et~al.(2023)Lai, Chen, Lu, Liu, and Jia]{Lai_2023_CVPR}
Xin Lai, Yukang Chen, Fanbin Lu, Jianhui Liu, and Jiaya Jia.
\newblock Spherical transformer for lidar-based 3d recognition.
\newblock In \emph{CVPR}, pages 17545--17555, 2023.

\bibitem[Lakshminarayanan et~al.(2017)Lakshminarayanan, Pritzel, and Blundell]{NIPS2017_9ef2ed4b}
Balaji Lakshminarayanan, Alexander Pritzel, and Charles Blundell.
\newblock Simple and scalable predictive uncertainty estimation using deep ensembles.
\newblock In \emph{NeurIPS}. Curran Associates, Inc., 2017.

\bibitem[Landrieu and Simonovsky(2018)]{Landrieu_2018_CVPR}
Loic Landrieu and Martin Simonovsky.
\newblock Large-scale point cloud semantic segmentation with superpoint graphs.
\newblock In \emph{CVPR}, 2018.

\bibitem[Lei et~al.(2019)Lei, Akhtar, and Mian]{Lei_2019_CVPR}
Huan Lei, Naveed Akhtar, and Ajmal Mian.
\newblock Octree guided cnn with spherical kernels for 3d point clouds.
\newblock In \emph{CVPR}, 2019.

\bibitem[Li et~al.(2019)Li, Muller, Thabet, and Ghanem]{Li_2019_ICCV}
Guohao Li, Matthias Muller, Ali Thabet, and Bernard Ghanem.
\newblock Deepgcns: Can gcns go as deep as cnns?
\newblock In \emph{ICCV}, 2019.

\bibitem[Li and Dong(2023)]{Li_2023_CVPR}
Jianan Li and Qiulei Dong.
\newblock Open-set semantic segmentation for point clouds via adversarial prototype framework.
\newblock In \emph{CVPR}, pages 9425--9434, 2023.

\bibitem[Li et~al.(2018)Li, Bu, Sun, Wu, Di, and Chen]{NEURIPS2018_f5f8590c}
Yangyan Li, Rui Bu, Mingchao Sun, Wei Wu, Xinhan Di, and Baoquan Chen.
\newblock Pointcnn: Convolution on x-transformed points.
\newblock In \emph{NeurIPS}. Curran Associates, Inc., 2018.

\bibitem[Li and Hoiem(2017)]{li2017learning}
Zhizhong Li and Derek Hoiem.
\newblock Learning without forgetting.
\newblock \emph{IEEE TPAMI}, 40\penalty0 (12):\penalty0 2935--2947, 2017.

\bibitem[Lis et~al.(2019)Lis, Nakka, Fua, and Salzmann]{Lis_2019_ICCV}
Krzysztof Lis, Krishna Nakka, Pascal Fua, and Mathieu Salzmann.
\newblock Detecting the unexpected via image resynthesis.
\newblock In \emph{ICCV}, 2019.

\bibitem[Meng et~al.(2019)Meng, Gao, Lai, and Manocha]{Meng_2019_ICCV}
Hsien-Yu Meng, Lin Gao, Yu-Kun Lai, and Dinesh Manocha.
\newblock Vv-net: Voxel vae net with group convolutions for point cloud segmentation.
\newblock In \emph{ICCV}, 2019.

\bibitem[Park et~al.(2022)Park, Jeong, Cho, and Park]{Park_2022_CVPR}
Chunghyun Park, Yoonwoo Jeong, Minsu Cho, and Jaesik Park.
\newblock Fast point transformer.
\newblock In \emph{CVPR}, pages 16949--16958, 2022.

\bibitem[Qi et~al.(2017{\natexlab{a}})Qi, Su, Mo, and Guibas]{Qi_2017_CVPR}
Charles~R. Qi, Hao Su, Kaichun Mo, and Leonidas~J. Guibas.
\newblock Pointnet: Deep learning on point sets for 3d classification and segmentation.
\newblock In \emph{CVPR}, 2017{\natexlab{a}}.

\bibitem[Qi et~al.(2017{\natexlab{b}})Qi, Yi, Su, and Guibas]{qi2017pointnet++}
Charles~Ruizhongtai Qi, Li Yi, Hao Su, and Leonidas~J Guibas.
\newblock Pointnet++: Deep hierarchical feature learning on point sets in a metric space.
\newblock In \emph{NeurIPS}, 2017{\natexlab{b}}.

\bibitem[Qian et~al.(2022)Qian, Li, Peng, Mai, Hammoud, Elhoseiny, and Ghanem]{NEURIPS2022_9318763d}
Guocheng Qian, Yuchen Li, Houwen Peng, Jinjie Mai, Hasan Hammoud, Mohamed Elhoseiny, and Bernard Ghanem.
\newblock Pointnext: Revisiting pointnet++ with improved training and scaling strategies.
\newblock In \emph{NeurIPS}, pages 23192--23204. Curran Associates, Inc., 2022.

\bibitem[Rethage et~al.(2018)Rethage, Wald, Sturm, Navab, and Tombari]{Rethage_2018_ECCV}
Dario Rethage, Johanna Wald, Jurgen Sturm, Nassir Navab, and Federico Tombari.
\newblock Fully-convolutional point networks for large-scale point clouds.
\newblock In \emph{ECCV}, 2018.

\bibitem[Riegler et~al.(2017)Riegler, Osman~Ulusoy, and Geiger]{Riegler_2017_CVPR}
Gernot Riegler, Ali Osman~Ulusoy, and Andreas Geiger.
\newblock Octnet: Learning deep 3d representations at high resolutions.
\newblock In \emph{CVPR}, 2017.

\bibitem[Tatarchenko et~al.(2018)Tatarchenko, Park, Koltun, and Zhou]{Tatarchenko_2018_CVPR}
Maxim Tatarchenko, Jaesik Park, Vladlen Koltun, and Qian-Yi Zhou.
\newblock Tangent convolutions for dense prediction in 3d.
\newblock In \emph{CVPR}, 2018.

\bibitem[Vaswani et~al.(2017)Vaswani, Shazeer, Parmar, Uszkoreit, Jones, Gomez, Kaiser, and Polosukhin]{NIPS2017_3f5ee243}
Ashish Vaswani, Noam Shazeer, Niki Parmar, Jakob Uszkoreit, Llion Jones, Aidan~N Gomez, \L~ukasz Kaiser, and Illia Polosukhin.
\newblock Attention is all you need.
\newblock In \emph{NeurIPS}. Curran Associates, Inc., 2017.

\bibitem[Wang et~al.(2018)Wang, Suo, Ma, Pokrovsky, and Urtasun]{Wang_2018_CVPR}
Shenlong Wang, Simon Suo, Wei-Chiu Ma, Andrei Pokrovsky, and Raquel Urtasun.
\newblock Deep parametric continuous convolutional neural networks.
\newblock In \emph{CVPR}, 2018.

\bibitem[Wang et~al.(2019)Wang, Sun, Liu, Sarma, Bronstein, and Solomon]{wang2019dynamic}
Yue Wang, Yongbin Sun, Ziwei Liu, Sanjay~E Sarma, Michael~M Bronstein, and Justin~M Solomon.
\newblock Dynamic graph cnn for learning on point clouds.
\newblock \emph{ACM Transactions on Graphics}, 38\penalty0 (5):\penalty0 1--12, 2019.

\bibitem[Wang et~al.(2021)Wang, Li, Che, Zhou, Liu, and Li]{Wang_2021_ICCV}
Yezhen Wang, Bo Li, Tong Che, Kaiyang Zhou, Ziwei Liu, and Dongsheng Li.
\newblock Energy-based open-world uncertainty modeling for confidence calibration.
\newblock In \emph{ICCV}, pages 9302--9311, 2021.

\bibitem[Wu et~al.(2019)Wu, Qi, and Fuxin]{Wu_2019_CVPR}
Wenxuan Wu, Zhongang Qi, and Li Fuxin.
\newblock Pointconv: Deep convolutional networks on 3d point clouds.
\newblock In \emph{CVPR}, 2019.

\bibitem[Wu et~al.(2022)Wu, Lao, Jiang, Liu, and Zhao]{NEURIPS2022_d78ece66}
Xiaoyang Wu, Yixing Lao, Li Jiang, Xihui Liu, and Hengshuang Zhao.
\newblock Point transformer v2: Grouped vector attention and partition-based pooling.
\newblock In \emph{NeurIPS}, pages 33330--33342. Curran Associates, Inc., 2022.

\bibitem[Xia et~al.(2020)Xia, Zhang, Liu, Shen, and Yuille]{10.1007/978-3-030-58452-8_9}
Yingda Xia, Yi Zhang, Fengze Liu, Wei Shen, and Alan~L. Yuille.
\newblock Synthesize then compare: Detecting failures and anomalies for semantic segmentation.
\newblock In \emph{ECCV}, pages 145--161, Cham, 2020. Springer International Publishing.

\bibitem[Xu et~al.(2020)Xu, Wu, Wang, Zhan, Vajda, Keutzer, and Tomizuka]{xu2020squeezesegv3}
Chenfeng Xu, Bichen Wu, Zining Wang, Wei Zhan, Peter Vajda, Kurt Keutzer, and Masayoshi Tomizuka.
\newblock Squeezesegv3: Spatially-adaptive convolution for efficient point-cloud segmentation.
\newblock In \emph{ECCV}, pages 1--19. Springer, 2020.

\bibitem[Zhang et~al.(2020)Zhang, Zhou, David, Yue, Xi, Gong, and Foroosh]{Zhang_2020_CVPR}
Yang Zhang, Zixiang Zhou, Philip David, Xiangyu Yue, Zerong Xi, Boqing Gong, and Hassan Foroosh.
\newblock Polarnet: An improved grid representation for online lidar point clouds semantic segmentation.
\newblock In \emph{CVPR}, 2020.

\bibitem[Zhao et~al.(2019)Zhao, Jiang, Fu, and Jia]{Zhao_2019_CVPR}
Hengshuang Zhao, Li Jiang, Chi-Wing Fu, and Jiaya Jia.
\newblock Pointweb: Enhancing local neighborhood features for point cloud processing.
\newblock In \emph{CVPR}, 2019.

\bibitem[Zhao et~al.(2021)Zhao, Jiang, Jia, Torr, and Koltun]{Zhao_2021_ICCV}
Hengshuang Zhao, Li Jiang, Jiaya Jia, Philip~H.S. Torr, and Vladlen Koltun.
\newblock Point transformer.
\newblock In \emph{ICCV}, pages 16259--16268, 2021.

\bibitem[Zhou et~al.(2021)Zhou, Ye, and Zhan]{Zhou_2021_CVPR}
Da-Wei Zhou, Han-Jia Ye, and De-Chuan Zhan.
\newblock Learning placeholders for open-set recognition.
\newblock In \emph{CVPR}, pages 4401--4410, 2021.

\end{thebibliography}
}


\end{document}